\documentclass{ecai}
\usepackage[T1]{fontenc}
\usepackage{graphicx}
\usepackage{latexsym}
\usepackage{soul}

\usepackage{microtype}
\usepackage{graphicx}
\usepackage{subfigure}
\usepackage{float}
\usepackage{caption}
\usepackage{booktabs} % for professional tables
\usepackage{amsmath}
\usepackage{amssymb}
\usepackage{mathtools}
\usepackage{amsthm}
\usepackage{algorithm}
\usepackage{algorithmic}
\usepackage{xcolor}
\usepackage{hyperref}
\usepackage{soul}
\usepackage{multicol}
\usepackage{multirow}
\usepackage{mdframed}
\theoremstyle{plain}

\begin{document}

%%%%%%%%%%%%%%%%%%%%%%%%%%%%%%%%
% THEOREMS
%%%%%%%%%%%%%%%%%%%%%%%%%%%%%%%%
\theoremstyle{plain}
\newtheorem{proposition}[theorem]{Proposition}
\newtheorem{lemma}[theorem]{Lemma}
\newtheorem{corollary}[theorem]{Corollary}
\theoremstyle{definition}
\newtheorem{definition}[theorem]{Definition}
\newtheorem{assumption}[theorem]{Assumption}
\theoremstyle{remark}
\newtheorem{remark}[theorem]{Remark}

%%%%%%%%%%%%%%%%%%%%%%%%%%%%%%%%%%%%

%%%%%%%%%%%%%%%%%%%%%%%%%%%%%%%%%%%%
% FRONT MATTER
\begin{frontmatter}

\title{Resource-constrained knowledge diffusion processes\\
inspired by human peer learning}

\author{\fnms{Ehsan}~\snm{Beikihassan}}
\author{\fnms{Amy K.}~\snm{Hoover}} % use of \orcid{} is optional
\author{\fnms{Ioannis}~\snm{Koutis} \thanks{E.~Beikihassan, A.K.~Hoover, and I.~Koutis contributed equally to this work. \\Corresponding author: \textit{ikoutis@njit.edu}}}
\author{\fnms{Ali}~\snm{Parviz}}
\author{\fnms{Niloofar}~\snm{Aghaieabiane} }

% AH \orcid{0000-0002-4661-8178}
% IK \orcid{0000-0003-1535-3397}
% NA \orcid{0000-0003-1096-7592}

\address{Ying Wu College of Computing\\ New Jersey Institute of Technology\\ 
Newark, NJ 07102, USA \\
\{eb283,~ahoover,~ikoutis,~ap2248,~na396\}@njit.edu }
%\author{\fnms{Ehsan}~\snm{Beikihassan}\thanks{Email: eb283@njit.edu.}}
%\author{\fnms{Yiannis}~\snm{Koutis}\thanks{Email: ikoutis@njit.edu.}}
%\author{\fnms{Amy K.}~\snm{Hoover}\thanks{Email: ahoover@njit.edu.}}
%\author{\fnms{Ali}~\snm{Parviz}\thanks{Email: ap2248@njit.edu.}}
%\author{\fnms{Niloofar}~\snm{Aghaieabiane}\thanks{Email: na396@njit.edu.}}

%\address{New Jersey Institute of Technology}

\begin{abstract}
We consider a setting where a population of artificial learners is given, and the objective is to optimize aggregate measures of performance, under constraints on training resources. The problem is motivated by the study of peer learning in human educational systems. In this context, we study  {\em natural} knowledge diffusion processes in networks of interacting artificial learners. By `natural', we mean processes that reflect human peer learning where the students' internal state and learning process is mostly opaque, and the main degree of freedom lies in the formation of peer learning groups by a coordinator who can potentially evaluate the learners before assigning them to peer groups. Among else, we empirically show that such processes indeed make effective use of the training resources, and enable the design of modular neural models that have the capacity to generalize without being prone to overfitting noisy labels.

\end{abstract}

\end{frontmatter}
%%%%%%%%%%%%%%%%%%%%%%%%%%%%%%%%%%%%

\section{Introduction}

A core issue in human educational systems is how best to use the existing resources, or more broadly how to diffuse knowledge over networks of human interactions that reflect pragmatic constraints~\cite{cowan_network_2004}. In particular, in educational settings, humans learn in peer groups. Recent literature has introduced quantitative models and knowledge diffusion processes for peer learning. While these models are simple, they appear to give insights that may have practical implications for peer learning~\cite{dong:dygroup}. This provides motivation for a study of analogous knowledge diffusion processes in the context of machine learning. 

In human educational systems, the objective is, broadly speaking, to train multiple individuals in finite time horizons and under resource constraints, e.g.~the scarcity of teachers and their bounded individual teaching capacity. In the context of machine learning, we can informally define a proxy problem as follows: We are given a  population of parametric models, a finite budget of accesses to the training set labels, and a finite time budget, and the objective is to maximize measures of average performance over the population.

\vspace{-.2cm}

\subsection{Natural Knowledge Diffusion}
\label{sec:KDiff}

Towards addressing the above  problem, we introduce our {\em natural Knowledge Diffusion} framework (\textsc{nKDiff}), that models human peer learning processes, following the paradigm  developed in~\cite{dong:dygroup}.

\begin{figure}[h] 
\begin{center}
    \includegraphics[scale=0.26]{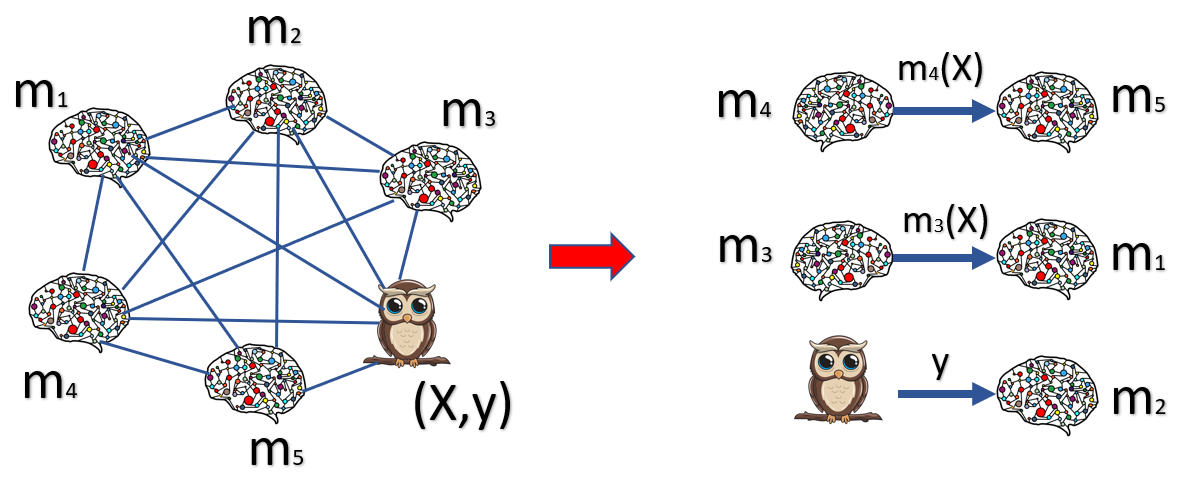}
\end{center}
\caption{Groups during one round of training} 
\label{fig:nets}
\end{figure}

\textsc{nKDiff} implicitly defines a set of admissible natural training processes for populations of learners.  As illustrated  in  Figure~\ref{fig:nets}, we are given a population of identical-architecture neural models that communicate with each other in {\em rounds}. We consider only the special case of a fully connected network of learners, indicating that all pairwise communications are possible. The network includes an {\em Oracle} agent that holds the true labels $y$. The training points $X$ are available to all models. The goal is to diffuse the Oracle's knowledge over the network and train all models with the objective of optimizing measures of {aggregate} performance. Towards the goal, we enable models to also operate as teachers by providing to other parts their {\em pseudolabels} $m_i(X)$.~\footnote{Note that participant models learn only through standard backpropagation training with labels they receive from other participants. Participants are oblivious of their own and the other participants' parameters, backpropagation gradients, and training losses, i.e.~their learning processes are handled by individual black-box optimizers.} We also impose the following constraints on the training resources of the process:

(i)~\textit{Training Capacity Constraint}: We set a bound on how many learners can interact with a teacher in each round of learning.

(ii)~\textit{Time Constraint}: We consider finite time budgets, typically smaller than what is required for `convergence' to a feasible optimal. In other words, we favor {\em fast diffusion}.

%This reflects the natural constraint that humans --and possibly humanity as a whole-- have: a finite learning horizon.%\footnote{Taking the metaphor to a poetic extreme: The goal of human education systems should be to improve the human population as fast as possible, before our planet resources run out so that we can escape Earth.} 

\subsection{Significant Observations}

The value of initiating and pursuing a study of natural peer learning processes in the context of machine learning is validated by the remarkable generalization properties of the various \textsc{nKDiff} processes we consider in this work (see Section~\ref{sec:peer}). \pagebreak

These are our main questions and significant observations:

\smallskip

\textbf{A.} {\em Are partially trained teachers detrimental or useful?}

It is not a priori clear to what extent employing partially trained teachers can be detrimental to population training. Surprisingly, the top accuracy reached by 
\textsc{nKDiff} processes {\em at convergence}, is comparable to that of baselines that do not employ partially trained teachers. 
When looking at finite time budgets (i.e.~in pre-convergent states), partially trained teachers are actually {\em useful}. \textsc{nKDiff} reaches higher accuracy significantly faster than standard population training algorithms with respect to the number of accesses to the training set. In other words, \textsc{nKDiff} makes more efficient use of the constrained training resources, which is precisely its main objective. In effect, \textsc{nKDiff} trades off accesses to the training set with parameters that are distributed over independent learners \footnote{For example, similar levels of test accuracy can be reached by using 9x parameters but accessing the training set 9x less frequently.}
(Section~\ref{sec:CEffect})

\medskip

\textbf{B.} {\em What is the impact of grouping policies?}

As elaborated in section~\ref{sec:formalization}, the main degree of freedom in \textsc{nKDiff} is in deciding the peer groups in each training round.  We find that the choice of a specific mechanism or policy for \textsc{nKDiff}  has a clear impact on performance measures, largely corroborating the results obtained via simple analytical models in~\cite{dong:dygroup}. (Sections~\ref{sec:policies}~and~\ref{sec:policyEffect})

\medskip

\textbf{C.} {\em Does training diversity have positive effects?}

A population of models trained with \textsc{nKdiff} consists of individuals that have undergone very diverse training processes due to their interactions with different teachers whose pseudolabels define constantly evolving loss functions. It is then natural to ask what the population-level effect of this diversity is. 
We find that when the population of learners is construed as an {\em ensemble} model, \textsc{nKdiff} 
prevents the ensemble from memorizing random training labels {\em despite} the individual capacity of its members to do so~\cite{zhang:rethinking}. We also find that the population can still generalize in noisy label settings, without overfitting to the noisy labels. Combining these observations we arrive at the conclusion that, as a training framework, \textsc{nKdiff} allows the composition of simple learning modules into a single model that has the capacity for generalization, but lacks the capacity for overfitting. (Section~\ref{sec:generalization})

\medskip
These and other observations point to various topics and previously observed phenomenal in machine learning that may offer partial justifications for \textsc{nKDiff} properties. We provide a related discussion in Section~\ref{sec:justification}

\vspace{-.3cm}

\subsection{Related Work}
\label{sec:related}

\textbf{Distributed Optimization.} The \textsc{nKDiff} algorithms we consider in this work can be viewed as distributed optimization algorithms. There is a large related literature on distributed optimization \cite{KonecnyMR15,SmithFMTJJ17} and federated learning~\cite{yangliu:federated/ml,ghoshChung:federated/Clustered}. Our work is more closely related to recent works on (decentralized) learning of personalized models through interactions of learners \cite{paul:decentralized,bellet:personalized/p2p,suiwen:personalized/Federated}. All these works ascribe algorithmic intent to the learners and consider algorithms for jointly optimizing them, thus requiring access to their internal state (e.g., gradients, or parameters) that may be exchanged with their neighbors on the network. They are also motivated by practical problems where the learners hold different data with true labels, so the main problem is to arrive at better personalized models (and/or a global model) that combine local data. In sharp contrast, \textsc{nKDiff} algorithms do not `open the box' of parameters, gradients, or training losses, and rely on oblivious learners that simply see and passively trust the pseudolabels provided by their designated teachers. In that sense, our work derives from the well-researched topic of information diffusion in social networks (e.g. see~\cite{cowan_network_2004,cnx019}. Many of these works on information diffusion fix simple mechanisms of information broadcasting and examine the impact of the network structure. On the other hand, the work on peer learning mechanisms in~\cite{dong:dygroup} fixes a simple network and considers the effect of broadcasting mechanisms. Our work is inspired by the latter paradigm.

\textbf{Ensemble Training.} A population of trained models can be naturally viewed as a single ensemble classifier. The focus of ensemble learning algorithms is on maximizing ensemble accuracy, and the number of the ensemble constitutes is a hyperparameter that frequently can be picked to be a relatively small number \cite{bonab:num-ensembles}. It is clear that a population of learners can have high ensemble accuracy with a low average accuracy; to see that, it suffices to think of a population containing only a few trained individuals while the rest return random outputs. Thus our objective of training a {\em given} number of learners for average accuracy is considerably different. Nevertheless, the time resource is also of interest in the context of ensemble learning as well. Boosting algorithms such as Adaboost~\cite{yoav:firtsboost,yoav:adaboost} and Gradient Boost~\cite{breiman:grad,friedman:grad} sequentially add classifiers to the ensemble over time, hence they are not time-aware. Bagging algorithms on the other hand train the ensemble's constituent models in parallel, improving accuracy under time constraints. Our results indicate that when viewed as ensemble training, knowledge diffusion has markedly different properties relative to plain bagging baselines. However, a deeper study in this direction is not in this paper's main scope.

\textbf{Education-inspired ML.} This paper adds to the broad literature on machine learning methods inspired by human education. We have in particular drawn inspiration from Curriculum Learning (CL)~\cite{bengio:curriculum,wu:curr-work,soviany:survey} and Knowledge Distillation~\cite{distilling:2015}. The main premise of CL is that {\em single} learners can reach higher levels of generalization (and possibly earlier in the training process) by carefully planning the way the training set is presented to the (passive) model undergoing the training. Related to CL is Active Learning (AL)~\cite{beluch:active-learning,burr:active} where control to generalization and training speed is achieved through the active choice of training points by a single learner. CL, and AL are somewhat related to our work, in the sense that they are resource-aware methods, but they aim at different objectives, they are orthogonal to our knowledge diffusion mechanisms and they can be even combined with them. We thus do not consider them further in this paper. 

\textbf{Empirical Phenomena in Deep Learning}.  Our work is closely related to research on empirical phenomena in deep learning \cite{sedghi:icmlws19}. In particular, in Section~\ref{sec:justification}, we discuss how some of our findings may be partly explained by phenomena related to the disagreement of randomly initialized models~\cite{jiang:iclr22, nakkiran:arxiv20} and the 
simplicity bias \cite{kalimeris:neurips19, arpit:icml17}. We are also inspired by the influential work of~\cite{zhang:rethinking} on generalization, and the subsequent work on learning from noisy labels~\cite{noisySurvey}. 
Notably, the idea of using a model's pseudolabels to train a learner has appeared in Co-Teaching~\cite{han_co-teaching_2018}, a training method involving two models. In Co-Teaching, each model takes into account training losses and based on them `cherry-picks' the training points/pseudolabels that are fed to the other model. In contrast, in our framework, the models do not have access to training/validation losses, and they are not even designed to be `aware' of when they interact with the Oracle, i.e. when they see true labels. Furthermore, teachers feed their pseudolabels indiscriminately to the learners. Thus, robustness to noise emerges as a byproduct of a natural process, unlike Co-Teaching and other training algorithms~\cite{noisySurvey} that are explicitly designed to deal with noisy labels. 

\subsection{\textsc{nKdiff}: Intuition and justification}
\label{sec:justification}

\textsc{nKDiff} mechanisms employ multiple partially trained teachers and one Oracle agent that holds the training set. These teachers provide pseudolabels to their students without checking the quality of the information they emit, hence diffusing false information. Moreover, each learner interacts with the Oracle for only a fraction of the time. Despite that, we find that partially trained teachers are useful as they enable a more efficient utilization of the constrained training resources. We view this phenomenon as a manifestation of overparameterization\footnote{By `overparameterization' we mean --somewhat loosely-- a number of parameters that suffices for reaching high training accuracy on noisy labels.} 
used in tandem with first-order training methods. In particular, we argue that `excess' parameters may be what causes the natural emergence of population-level learning, which in turn naturally prevents overfitting to noisy labels.

\textit{Overparameterization for diverse learners.} Large models can be pruned to sparse `lottery' models that can reach or exceed the test performance of the original dense models~\cite{franckle:LotteryHypothesis}. Finding these sparse models requires elaborate initialization or pruning techniques of the fully dense models~\cite{kartik:RareGems}. In other words, sparsity makes initialization a very delicate issue. 

On the flip side, a remarkable 
empirically observed property of (dense) models trained with gradient descent is that two randomly initialized models tend to have very similar test accuracy, but they also have a high rate of disagreement which is almost equal to their test error~\cite{jiang:iclr22,nakkiran:iclr20}.

\begin{figure}[h!]
    \centering
    {\includegraphics[scale=0.1]{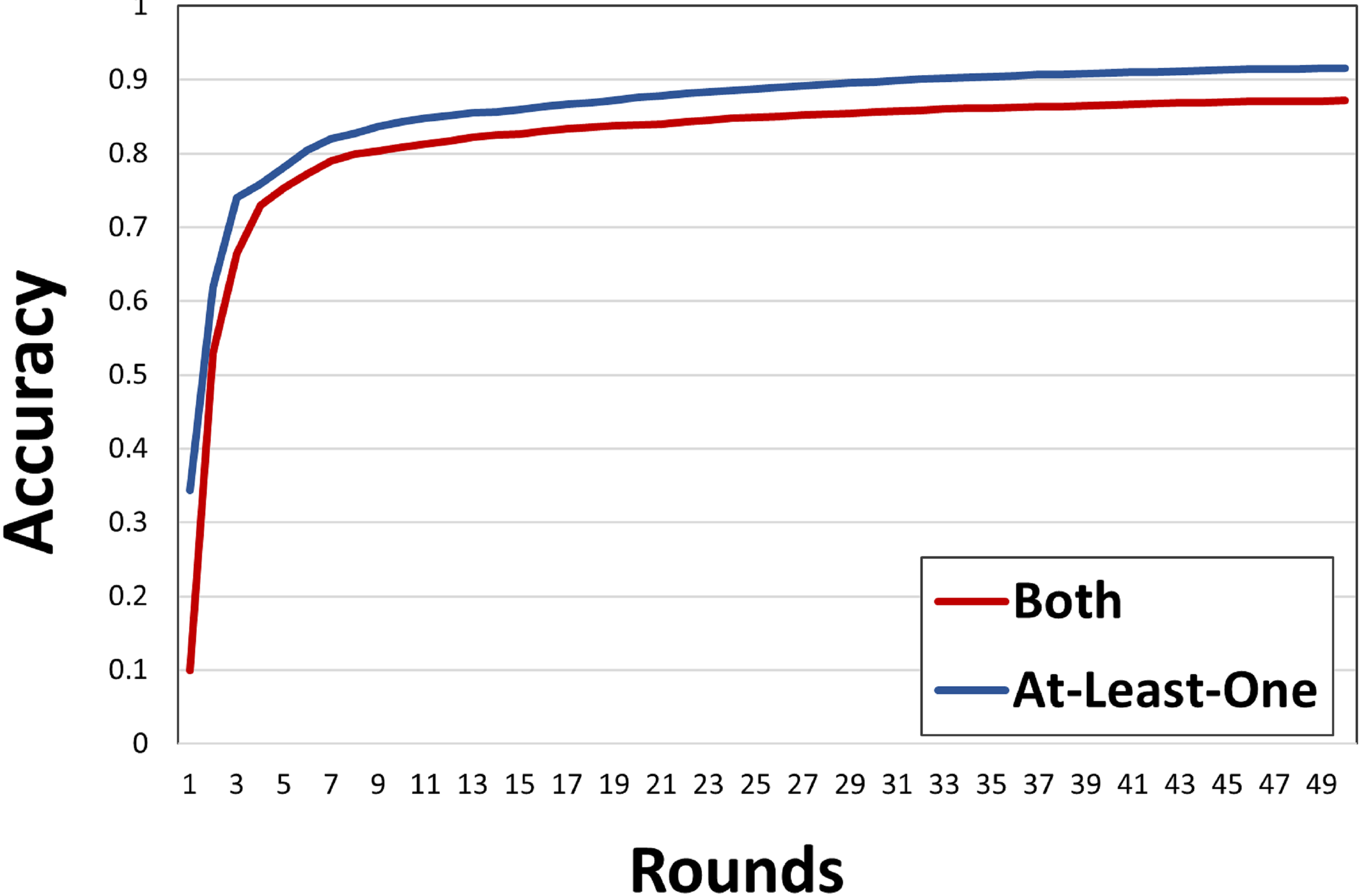}}
    \caption{Training two randomly initialized LeNet models  and tracking the number of test points that are correctly classified by {\em both} models vs the number of test points that are correctly classified by {\em at least one} model. The difference is more pronounced on bigger ResNet models.}
    \label{fig:hard-soft}
\end{figure}

Figure~\ref{fig:hard-soft} depicts an instance of this phenomenon, over the entire training period. The disagreement is captured by the gap between the two curves. We observe that the relative disagreement of the two models tends to be higher in the beginning of training when, according to previous works, the networks learn simpler functions~\cite{kalimeris:neurips19}. Intuitively this hints at a \textit{spatial lottery}: the two models learn different parts of the data `manifold' at different rates, due to randomness in initialization. Then, when one model is used as a teacher to the other, the student model can still learn information that it has not previously learned. In other words, the excess parameters not only make random initialization much easier, but render it a natural resource.

\textit{Learner diversity prevents `unlearning'.} In the presence of noisy labels, single models first undergo a phase of learning and reach high test accuracy. That is followed by an `unlearning' phase of overfitting to the noisy labels which negatively impacts test performance~\cite{arpit:icml17}. In the case of \textsc{nKDiff} all learners undergo the first phase and reach high test accuracy. However, the unlearning phase does not take place because the learners interact with the (noisy) Oracle only a fraction of the time, while seeing more consistent labels from their peers.

%%%%%%%%%%%%% PROBLEM FORMULATION

\section{\textsc{nKDiff}: Formulation and Mechanisms}
\label{sec:peer}

This section details and justifies the various \textsc{nKDiff} mechanisms we consider in this paper.

\subsection{Definitions and Problem Formulation}
\label{sec:formalization}

\textbf{Learners, Trainers and the Oracle Model.} We are given a {\em population} of $N$-1 classifier models, and we want to train them with a training set $X$ with categorical labels $y$. The models are artificial neural networks (ANNs) that have identical architectures. They are trained in {\em rounds}. 

\begin{figure} [h]
\begin{center}
 \includegraphics[scale = 0.3]{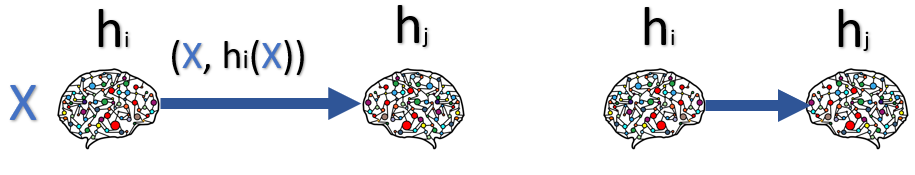}
     \caption{(Left:) Model $h_i$ acts as teacher to model $h_j$ by providing its predictions on the training dataset $X$ as labels for a training sessions with $h_j$. (Right:) A schematic abbreviation.} \label{fig:session}
\end{center}     
\end{figure}

During a round of training, each model acts as a learner or a teacher. When acting as a {\em teacher} a model provides its (partially correct) predictions $h_i(X)$ for the training set $X$ and it does not update its parameters. When acting as a {\em learner}, a model undergoes training with the training set provided by its teacher, and accordingly updates its parameters through standard iterations of forward and back-propagation operations. This is illustrated in Figure~\ref{fig:session}.

The {\em Oracle model} is the $N^{th}$ model, $h_N$; it always teaches the correct labels $y=h_N(X)$ when queried, and consequently ignores learning the predictions of other models.

\smallskip

\textbf{Prior Knowledge and Learner Initialization Schemes.} In an educational system, at any point of time, there are students that have different degrees of prior exposure to knowledge. Teachers also have varying levels of expertise. For that reason we allow the models to start the peer learning process after undergoing partial pre-training with the true labels. The type and amount of pre-training can be viewed as a hyperparameter of the peer learning mechanism. We call that the  {{\em learner initialization scheme}}. Of course, the case of no prior knowledge is also of interest. In such case the \textit{N-1} models are randomly initialized.

\textbf{Coordinator and Groups.} Training the \textit{N-1} classifiers takes place in rounds and it may include a {\em Coordinator}. The Coordinator implements {\em grouping policies}. More specifically, before each round the Coordinator can: (i) Evaluate the performance of the \textit{N-1} models (ii) Define groups of models and designate a single teacher for each group, for the upcoming round.  Any algorithmic process that determines these groups is considered a grouping policy.

\textbf{Learning Sessions.}  During a round, in each group, the teacher has an independent {\em session} with each of its assigned learners. In this paper, the session between a learner and teacher-model $h_i$ consists of an epoch over the entire training data $X$ with labels $h_i(X)$.

\begin{figure}[h] 
\begin{center}
\includegraphics[scale=0.28]{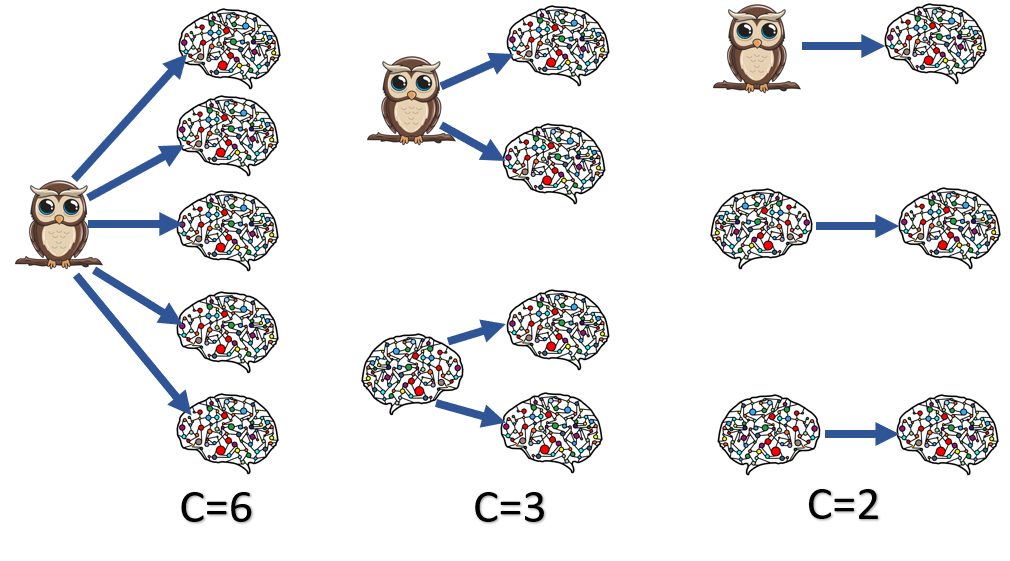}
\end{center}
\caption{Without a teaching capacity constraint ($C$=6), the Oracle can teach all learners in each round, defaulting to plain ensemble training. In our framework, any model can act as a teacher, and our convention is to consider teaching groups of equal size; here ($C$=3) and ($C$=2) are depicted.}
\label{fig:groupings}
\end{figure}

\textbf{Training Capacity Constraint.} We impose a {capacity bound} $C$ on the size of the groups in each round of training. That implies that any teacher can teach up to $C$-1 learners per round.  We will also be denoting by $k$ the number of groups per round. An illustration of training capacity constraints can be seen in Figure~\ref{fig:groupings}.

\textbf{Performance Metrics.}  In our setting we are typically interested in general `educational welfare' objectives~\cite{dong:dygroup}. For that reason, we use metrics of aggregate generalization performance, e.g.~the average test accuracy of the learners. However, a trained population can naturally be viewed as an ensemble, and thus ensemble test accuracy remains of interest in our context as well. To make things more concrete we introduce some definitions.

\begin{definition} [Average Learner Accuracy]
Let $\cal E$ be an ensemble of $m$ classifiers, $h_1,\ldots,h_m$. We define the average learner accuracy of $\cal E$ on a dataset $(X,y)$ by
$$
     alacc_{\cal E}(X,y) = \left(\sum_{i=1}^{m} acc_{h_i }(X,y)\right)/m
$$
\end{definition}

\begin{definition} [Ensemble Output]
Let $\cal E$ be an ensemble of identical-architecture classifiers, and further assume that the output of each classifier $h_i \in {\cal E}$ on an input point $x$ is a probability distribution $p_i\in {\mathbf R}^K$, where $p_i[j]$ denotes the probability assigned to class $j$.  The classification of $x$ by $\cal E$ is given by:
$$
  {\cal E}(x) = argmax_{j} \sum_{i} \log p_{i}[j].
$$
\end{definition}

\begin{definition} [Ensemble Accuracy]
The accuracy of a classifier $\cal C$ over a dataset $(X,y)$, denoted by $acc_{\cal C}(X,y)$, is the ratio of points in $X$ that are classified correctly by $\cal C$. Viewing $\cal E$ as a classifier, we refer to $acc_{\cal E}(X,y)$ as the ensemble accuracy. 
\end{definition}

\subsection{Peer learning group policies}
\label{sec:policies}
% \Yiannis{3. Section Done.}

In this section we review the policies included in our empirical study and provide some additional background and justification for them.

\begin{figure}[h]
\begin{center}
\includegraphics[scale=0.25]{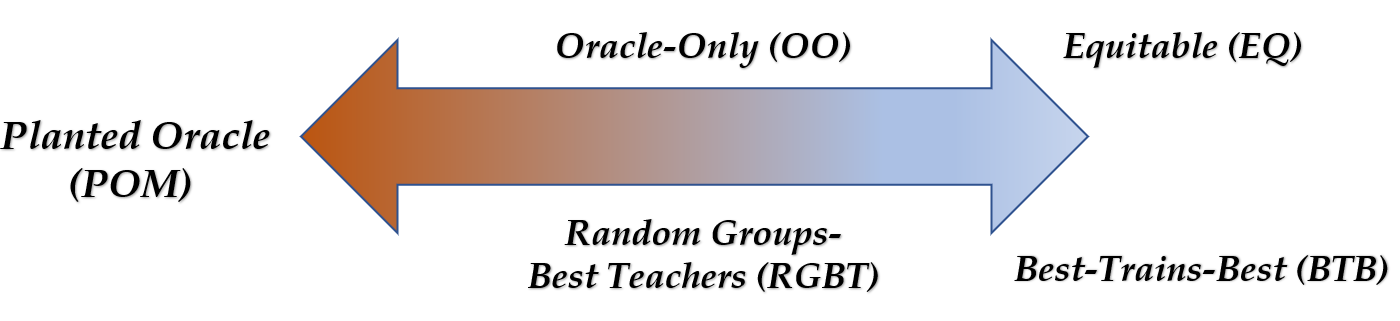}
\end{center}
\caption{{\bf Grouping policies in the spectrum of coordination}. \\POM is a decentralized and uncoordinated mechanism where the Oracle is concealed as a participant. OO and RGBT are moderately coordinated. In particular OO is a baseline that does not make use of partially trained teachers. EQ and BTB are fully coordinated policies aiming at different types of `social objectives'. }
\label{fig:spectrum}
\end{figure}

Previous works on analytical models for peer human learning consider a number of grouping policies that reflect a long-standing debate on class formation in the realm of the social sciences and education policy making~ (e.g. see~\cite{dominick:homogen,stephen:ability,jo:ability, asma:group-formation}).%~\footnote{The debate remains current and sensitive. In a well-publicized case, gifted and talented programs in New York City schools where to phased out in 2021, only to be revived in 2022 after an administration change.}  
Underlying the debate is the common-ground intuition that public policies can have an impact on the educational welfare of the population. In our study we pursue an `artificialization' of this question and consider a spectrum of policies with various degrees of coordination and underlying social intent, as illustrated in Figure~\ref{fig:spectrum}. More specifically, we consider these policies: 
%We wish to emphasize though that it is not our intention to take a stance or make policy recommendations. 

\smallskip

{\it \textbf{OO:}} {\em Oracle-Only}. In each round, \textit{C-1} models are selected at random and they are trained by the Oracle. We view this policy as a special case of our framework, and we include it as a baseline. 

\smallskip

{\it \textbf{POM:}} {\em Planted Oracle Mechanism.} The POM is our only fully decentralized policy, testing whether knowledge can be diffused without any evaluation of the learners or any other type of external coordination. In each round of the training process, the $N$ models are randomly split into $N/2$ pairs. Suppose that models $h_i$ and $h_j$ form a pair. Then two sessions take place, one where $h_i$ is trained with $(X,h_j(X))$ and one where $h_j$ is trained with $(X, h_i(X))$. In this case, the only possible training capacity is \textit{C=2}. Observe that this mechanism does not evaluate the models at any point of the process and that the Oracle model conceals itself as a participant in the training process.

\smallskip

The remaining grouping policies are based on measuring the validation accuracy of the learners  $h_1,\ldots,h_{N-1}$ before each round. These accuracies are communicated to the coordinator who then decides the grouping for the next round. While communication complexity is not our main concern, we note that if the learners hold copies of the validation set, then the complexity of this process is small; the coordinator needs to only receive a single number (validation accuracy) from each learner, and send them back the identity of their teacher for the next round. In what follows, we let $v_i$ be the validation accuracy of $h_i$, and $m_i$ denote the model whose validation accuracy is the $i^{th}$ lowest in the list   $\{v_1,\ldots,v_N\}$. 

\smallskip

{\it \textbf{RGBT:}} {\em Random-Groups Best-Teachers.} The models are randomly split into \textit{k= N/C} groups. Then the model with the highest validation accuracy in its group is designated as the teacher. In this approach, there is no coordination in the selection of groups, but only in the selection of teachers.

\smallskip

{\it \textbf{BTB:}} {\em Best-Trains-Best.} The policy uses the current-round best learners as teachers. It furthermore greedily assigns better students to better teachers. The trainers of the $k$ groups are models $m_N,\ldots,m_{N-k+1}$, i.e. the models with the highest validation accuracy. The rest of the ordered list $m_{N-k},\ldots, m_1$ is split into $k-1$ contiguous buckets that are assigned in order to $m_N,\ldots,m_{N-k+1}$. This corresponds to situations where better students are
grouped together in classes, and they learn from better
teachers.

\smallskip

{\it \textbf{EQ:}} \textit{Equitable.} The policy uses the current-round best learners as teachers. It furthermore greedily assigns students in order to create ‘balanced’ groups, in terms of the students’ ability. The policy attempts to
be fair (some weak students will be assigned to good teachers), while
still giving a slight edge to better students (e.g., the weakest student
will be matched with the weakest teacher). More concretely, the trainers of the $k$ groups are models $m_N,\ldots,m_{N-k+1}$. The rest of the models in the ordered list $m_{N-k},\ldots, m_1$ are assigned in a round-robin fashion to $m_N,\ldots,m_{N-k+1}$. In this grouping each teacher is assigned to learners at all levels of accuracy.

\section{The empirical study: Questions and Findings}
\label{sec:results}

In this section we present the main findings of our extensive empirical study. After reviewing our experimental setting in section~\ref{sec:setting}, we discuss the effectiveness of~\textsc{nKDiff} in section~\ref{sec:CEffect}, the performance difference among policies in section~\ref{sec:policyEffect}, and the generalization benefits of learner diversity in section~\ref{sec:generalization}.

\vspace{-.3cm}

\subsection{Experimental setting}
\label{sec:setting}
\textbf{Architectures and Datasets.} We perform experiments with three different types of architectures and two different types of datasets. In particular, we use `toy' versions of LeNet (5 layers with \textit{CrossEntropyLoss} function and \textit{Stochastic Gradient Descent} optimizer with learning rate of 0.9, where we have a total number of 61.7K parameters) and ResNet (18 layers with \textit{CrossEntropyLoss} function and \textit{Adam} optimizer with learning rate of 3e-4, where we have a total number of 11.1M parameters) ~\cite{yann:lenet,kaiming:resnet}. These networks are used on Fashion-MNIST dataset which consists of 50,000 training, 10,000 validation, and 10,000 test images of fashion and clothing items, taken from 10 classes, where each image is a standardized 28×28 size in grayscale. We also use a Graph Convolutional Network (GCN) (consisting of 3 layers, 32 hidden channels and dropout of 0.5, with \textit{CrossEntropyLoss} function, \textit{Adam} optimizer with learning rate of 0.01, where we have a total number of 8.9K parameters)~\cite{kipf:gcn,zhou:gcn}. This is employed for a transductive classification problem. We use the ogbn-arxiv dataset which is a un-directed graph representing the citation network between all Computer Science (CS) arXiv papers, where each node is an paper with a 128-dimensional feature vector which consists of 90K training, 48K validation and 29K testing points. 

%All experiments were conducted on Google Colab. 

\noindent \textbf{Peer Framework Settings.} In our study, we look at populations of size \textit{N=10}. We explored two settings for $C$, groups of size two (\textit{C=2}) and groups of size five (\textit{C=5}). Thus, we have $k=5$ and $k=2$ groups respectively. We refer to these settings as split-in-five and split-in-two. We include experiments without pre-training and with pre-training. In our learner initialization scheme, model $i$ has been trained for $i$ rounds with the true labels.

\noindent \textbf{Number of Random Experiments.} In each experiment we first randomly initialize each model. Then on the {\em same} initialized population, we try each combination (\textit{Network}, \textit{Pretraining/No-pretraining}, \textit{C}, \textit{Policy}). In the experiments of  Sections~\ref{sec:CEffect} and~\ref{sec:policyEffect} we
report {\em averages} of our metrics, taken over {100} random experiments. In the experiments of Section~\ref{sec:generalization} we report averages over the following numbers of random experiments: 10 for LeNet, 20 for ResNet, and 10 for GCN.  The number of random experiments has been picked in order to derive 95\% confidence intervals, shown in cases when the difference among policies was not extremely clear cut. Confidence levels are calculated in a standard way.~\footnote{We believe that actual confidence levels are much tighter than reported, but we stray away from such statistical arguments.}  

\smallskip
\noindent \textbf{Justification.} 
For the image classification experiments we fix a single dataset, which is difficult for the LeNet model, but relatively easy for the much larger and fundamentally more expressive ResNet model. With this choice we want to test how \textsc{nKDiff} works in different `hardness' and parameterization regimes, especially under the light of the discussion in Section~\ref{sec:justification} that identifies overparameterization and model disagreement as a cause underlying the effectiveness of \textsc{nKDiff}.  The GCN experiment is meant to test   \textsc{nKDiff} in a transductive setting, with a fundamentally different type of problem and architecture, and relatively fewer trainable parameters. 

\smallskip
\noindent \textbf{Code.} The code can be accessed here: \\
\url{https://github.com/peer-ai-njit/ecai23}
%\url{http://tiny.cc/peerAI}.

\subsection{The effectiveness of \textsc{nKDiff}}
\label{sec:CEffect}

$\bullet$ \textit{The price of teacher scarcity at convergence.} We take a look at the highest levels of performance the learners were able to reach without any time constraint.~\footnote{In this experiment ResNet was trained for 50 rounds, and LeNet and GCN were trained for 100 rounds. There was no significant change in their validation accuracy in the last 20\% of their epochs, so we consider them converged.} 
The results are shown in Table~\ref{fig:table-1}; here we use the BTB policy which performed best among \textsc{nKDiff} policies.

\renewcommand{\arraystretch}{1.2}
\begin{table}
\centering
\begin{tabular}{ |c|c|c|c|c|c|  }
\cline{3-5}
\multicolumn{1}{c}{}&
\multicolumn{1}{c}{}&
\multicolumn{1}{|c|}{C = 10} & \multicolumn{1}{c|}{C = 5} & \multicolumn{1}{c|}{C = 2} \\
\hline
\multirow{ 2 }{*}{ResNet} & \textit{Avg} & 90.4 $\pm$ 0.002     & 90.4 $\pm$ 0.001     & 90.4 $\pm$ 0.002    \\\cline{2-5}
& \textit{Ens}     & 91.6 $\pm$ 0.004     & 91.0 $\pm$ 0.004     & 90.1 $\pm$ 0.007    \\\cline{2-5}\hline
\multirow{ 2}{*}{LeNet} & \textit{Avg} & 89.5 $\pm$ 0.002     & 89.3 $\pm$ 0.003     & 86.9 $\pm$ 0.001    \\\cline{2-5}
& \textit{Ens}     & 89.8 $\pm$ 0.003     & 89.3 $\pm$ 0.003     & 86.6 $\pm$ 0.002    \\\cline{2-5}\hline
\multirow{ 2}{*}{GCN} & \textit{Avg} & 69.1 $\pm$ 0.003     & 67.4 $\pm$ 0.002     & 64.7 $\pm$  0.005   \\\cline{2-5}
& \textit{Ens}     & 69.7 $\pm$ 0.004     & 68.1 $\pm$ 0.004     & 67.4 $\pm$ 0.004    \\\hline

\end{tabular}
\caption{Impact of training capacity on the maximum performance attained by the population. \textit{Avg} denotes average test accuracy and \textit{Ens} ensemble test accuracy. The case \textit{C=10} is when all models are trained in parallel by the Oracle model, as in Figure~\ref{fig:groupings}(a). Standard deviation is over multiple random experiments (see Section~\ref{sec:setting}). \label{fig:table-1}}
\end{table}

When partially trained teachers are used i.e.~when \textit{C=2} and \textit{C=5}, the Oracle teacher is respectively accessed only $55.5\%$ and $11.1\%$ of the times accessed relative to \textit{C=10}, and its true labels are replaced by false/inconsistent labels. We thus expect to see an impact in test performance. We find though that this impact is relatively small. A potentially interesting fact is that the smallest impact is observed for ResNet, the most `overparameterized' of these architectures. 

\smallskip

\begin{figure}[h!]
     \centering
     \begin{subfigure}
         \centering
         \includegraphics[width=42mm]{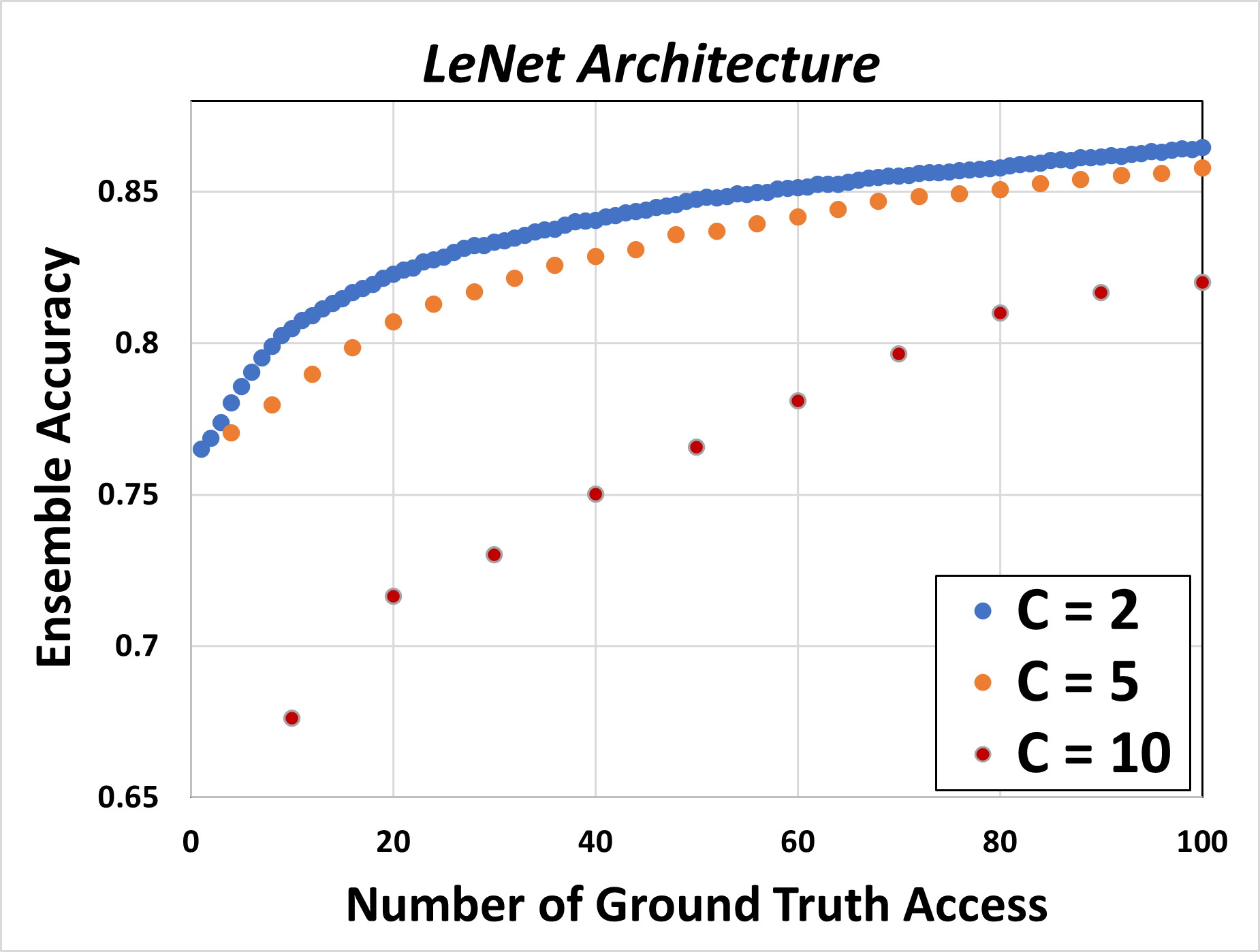}
     \end{subfigure}
     \begin{subfigure}
         \centering
         \includegraphics[width=42mm]{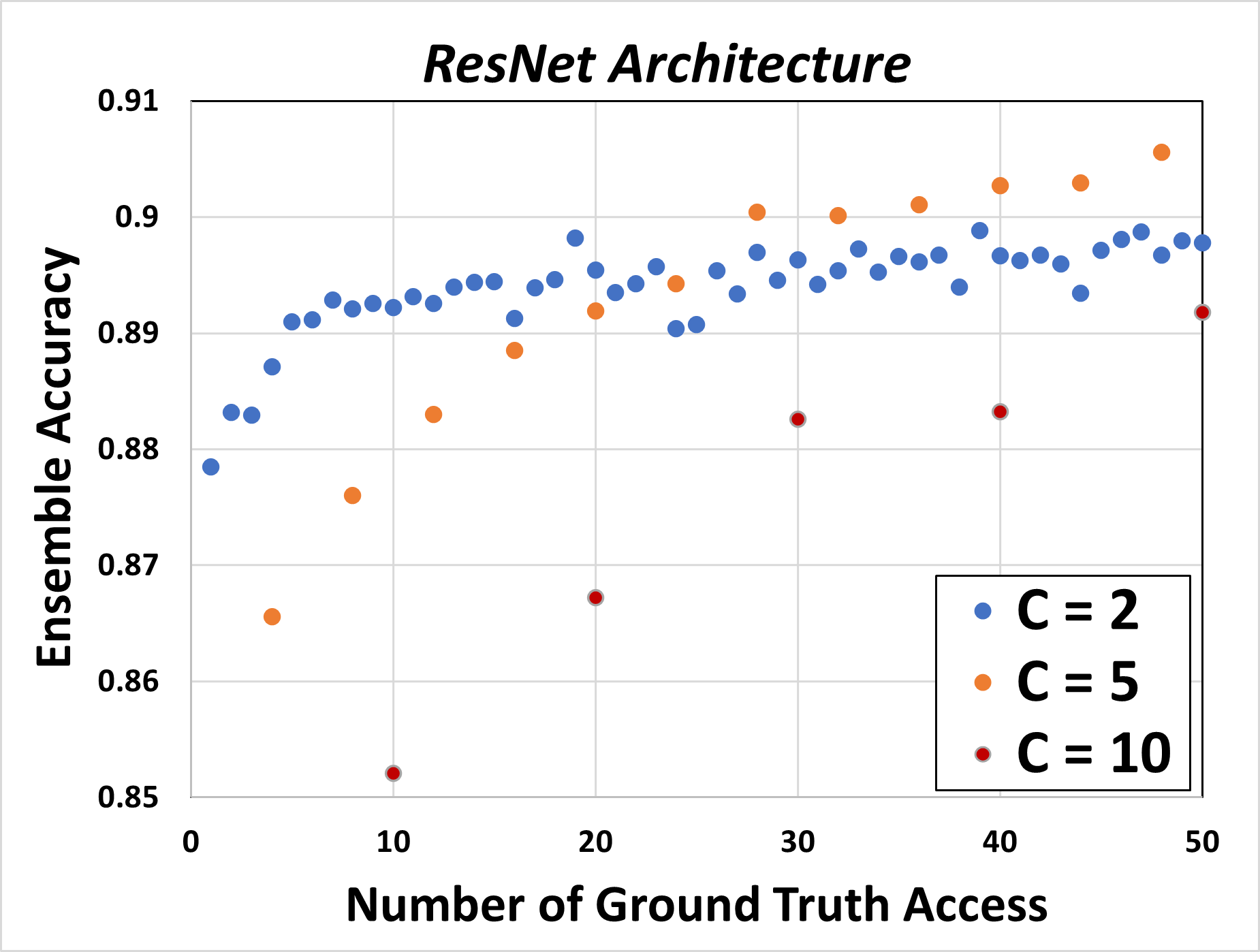}
     \end{subfigure}     
        \caption{Ensemble accuracy as a function of the number of oracle sessions, for \textit{C=2}, \textit{C=5} and \textit{C=10}. GCN behavior is qualitatively similar to LeNet. Also all other policies have similar behavior to {\em BTB}. }
        \label{fig:utilization}
\end{figure}

\noindent $\bullet$ {\em The usefulness of partially trained teachers.} Recall that the primary reason behind our \textsc{nKDiff} study is the efficient utilization of the teaching resources. We thus want to consider the test performance of the population as a function of the number of accesses of the true labels. The results are shown in Figure~\ref{fig:utilization}.\footnote{Results are similar for GCNs. Average learner accuracy as a function of the number of ground truth accesses is also similar to that of Figure~\ref{fig:utilization}. The plots can be found in the Supplementary Material.}.

We observe that a {\em smaller} training capacity $C$ leads to {\em higher} performance, for any given small budget of Oracle sessions. This implies that partially trained teachers are indeed helping in extracting knowledge from the Oracle more efficiently, and by a large margin in pre-convergent epochs! We also find that {\em smaller} classes (\textit{C=2}) are better. Interestingly, smaller classes employ more and thus weaker teachers. Overall the `natural' necessity of using more teachers, in tandem with overparameterization at the population level, lead to faster mining of the ground truth.

\pagebreak

\begin{figure}[h]
     \centering
     \begin{subfigure}
         \centering
         \includegraphics[width=42mm]{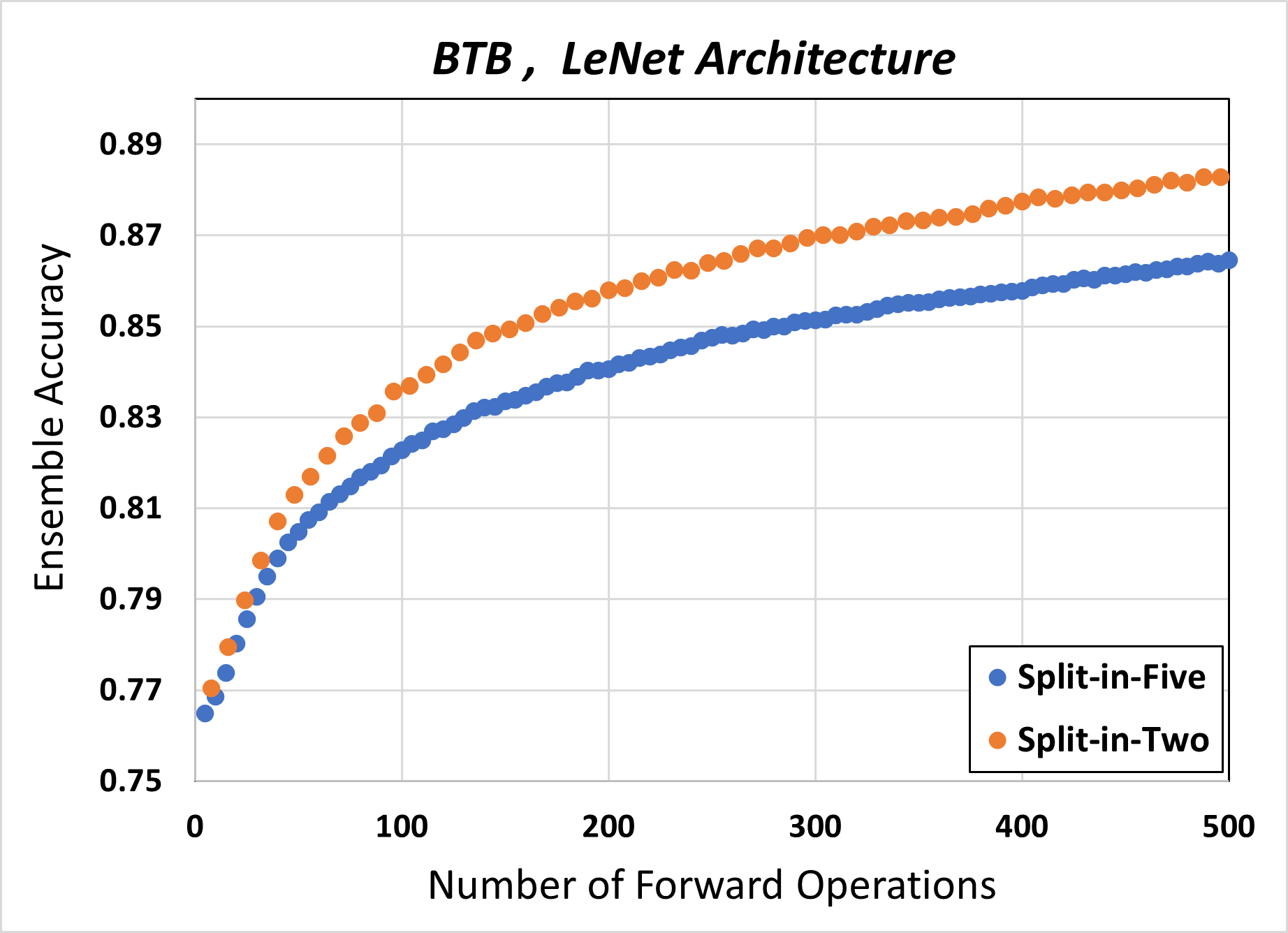}
     \end{subfigure}     
     \begin{subfigure}
         \centering
        \includegraphics[width=42mm]{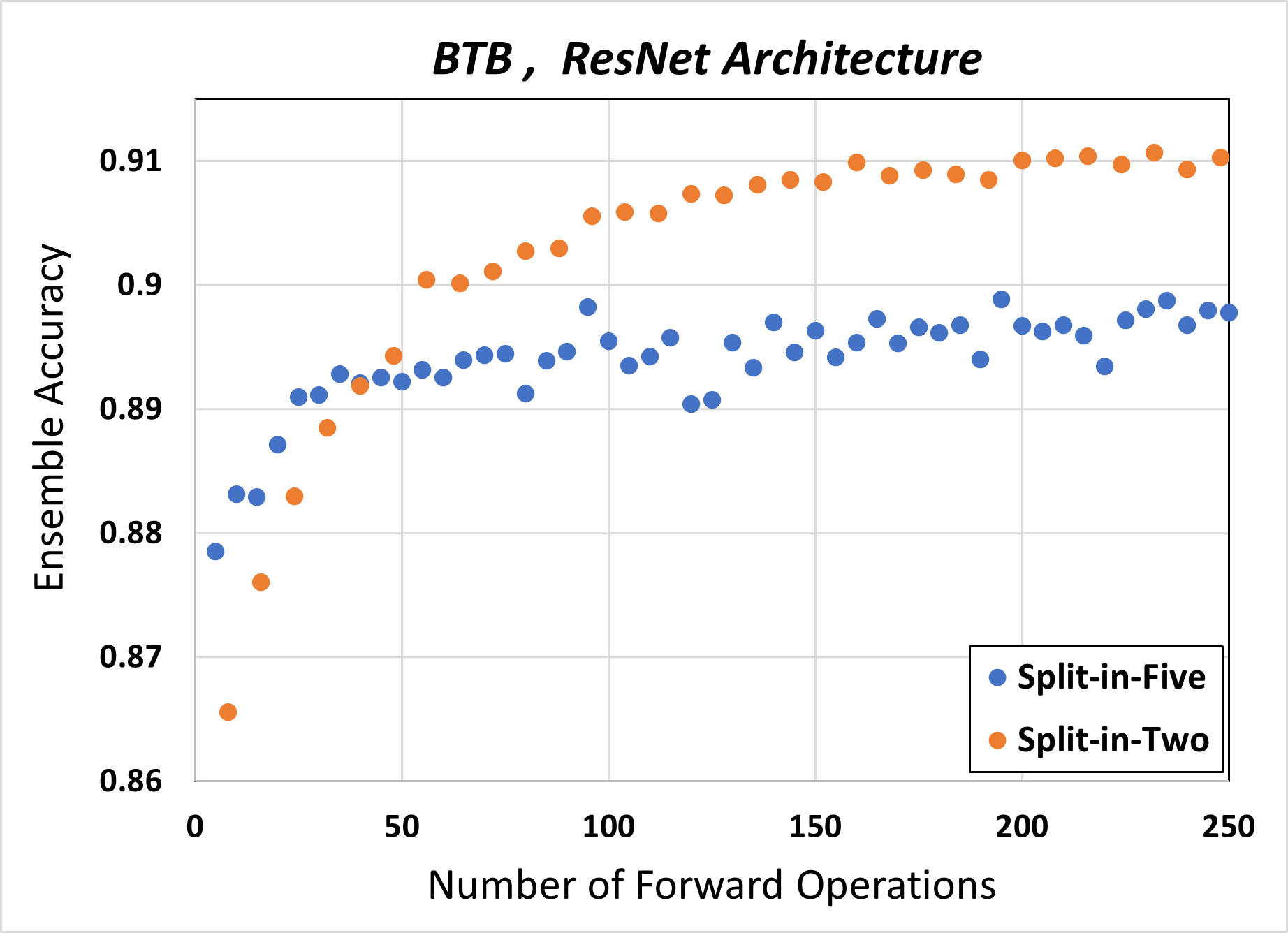}
     \end{subfigure}
    \caption{Ensemble test accuracy $acc_{\cal E}$ for LeNet and ResNet as a function of number of \textit{Forward Operations}.}
    \label{fig:forward-lenet-resnet}
\end{figure}

\noindent $\bullet$ {\em The economics of class size.} In an educational setting, employing weaker teachers is, of course, not free. We thus study an alternative cost model where we measure performance as a function of the number of training sessions (or equivalently, forward operations). 
The result is shown in Figure~\ref{fig:forward-lenet-resnet}.

Here we see larger classes (\textit{C=5}) have a higher knowledge extraction rate relative to the smaller classes (\textit{C=2}), which is not unexpected given the fact that larger classes employ better teachers. 

\subsection{Policy Effect}
\label{sec:policyEffect}

\noindent $\bullet$ {\it The Planted Oracle Mechanism.}  
In the \textit{POM}, all models act both as teachers and learners in every round. Models are not evaluated at any point of the training process, and the Oracle model conceals itself as a participant in the process. The Coordinator has the very limited role of simply timing the rounds. Thus POM corresponds to an extreme case of a completely unorganized population, with random interactions among its members who just exchange information. 

\begin{figure}[h]
     \centering
     \begin{subfigure}
         \centering
         \includegraphics[width=42mm]{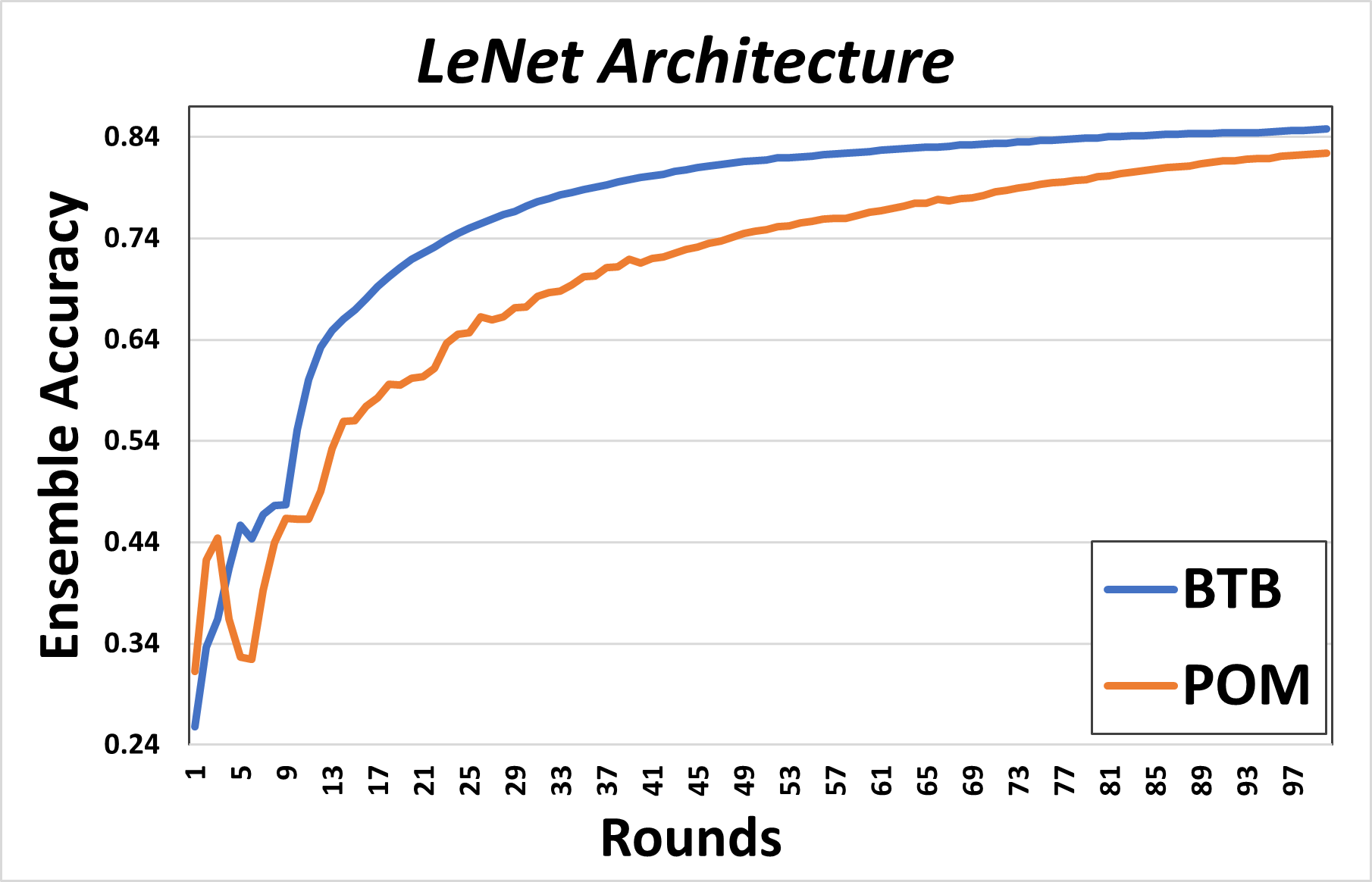}
     \end{subfigure}     
     \begin{subfigure}
         \centering
         \includegraphics[width=42mm]{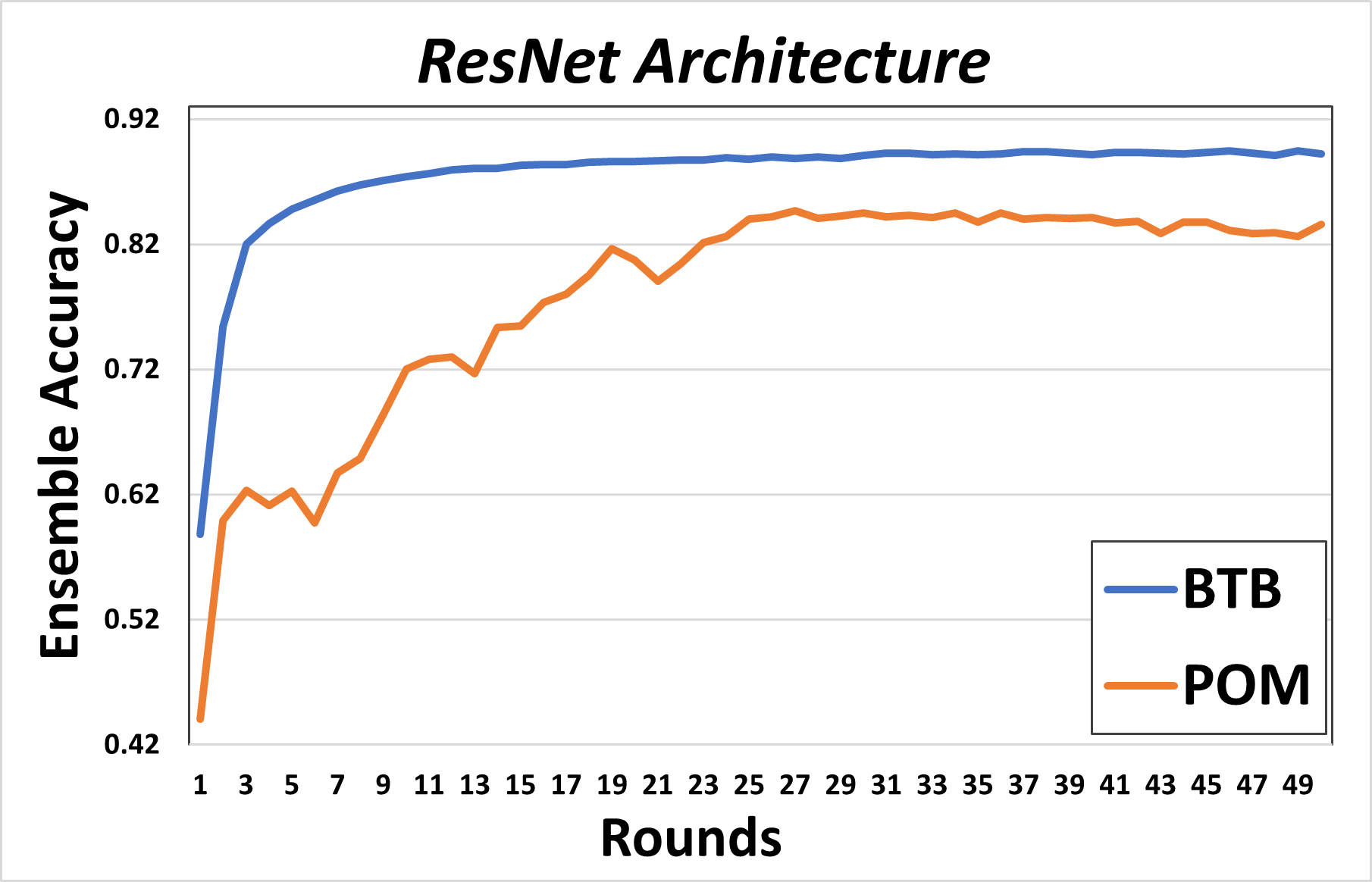}
     \end{subfigure}
     \begin{subfigure}
         \centering
         \includegraphics[width=42mm]{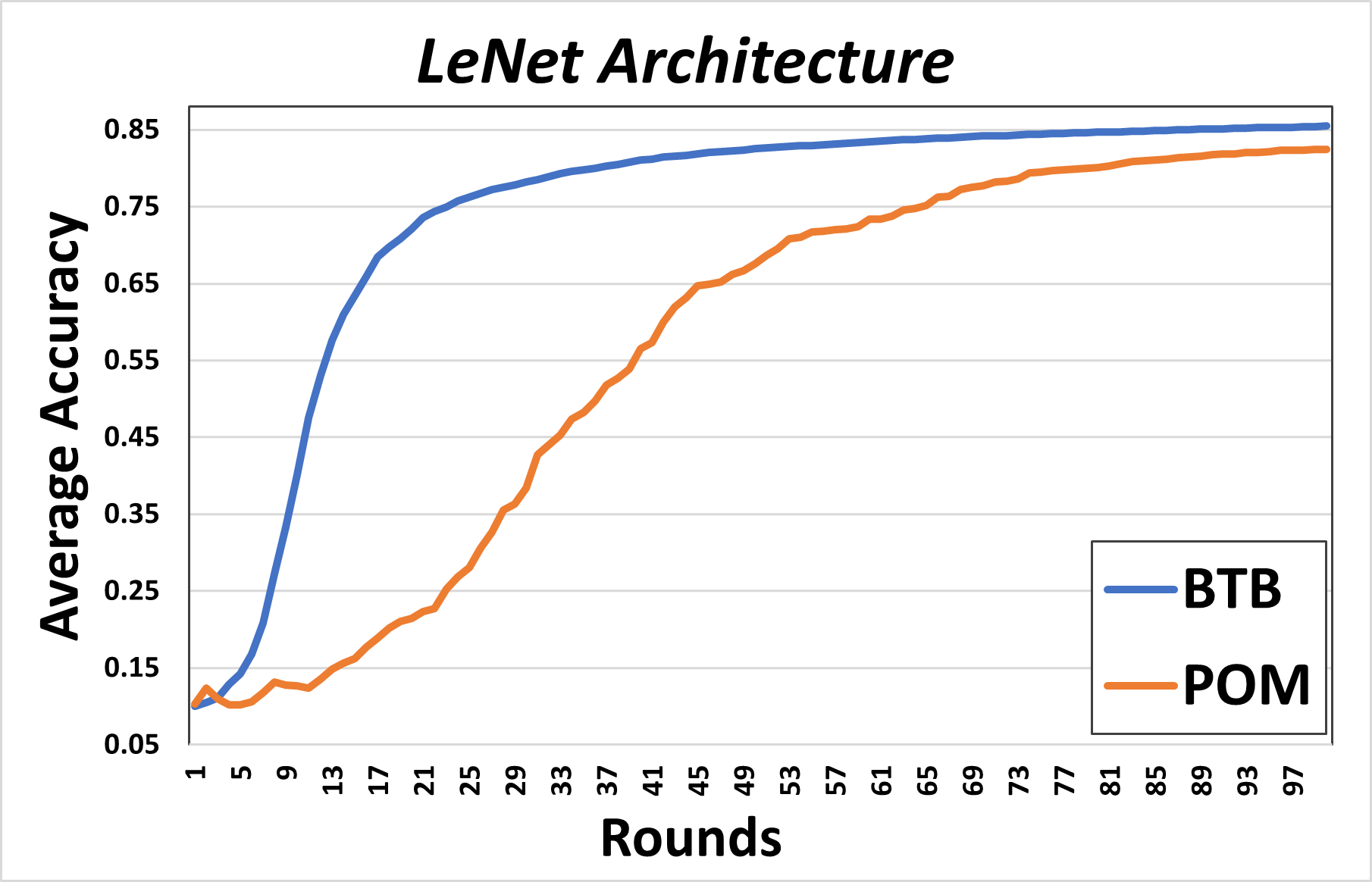}
     \end{subfigure}     
     \begin{subfigure}
         \centering
         \includegraphics[width=42mm]{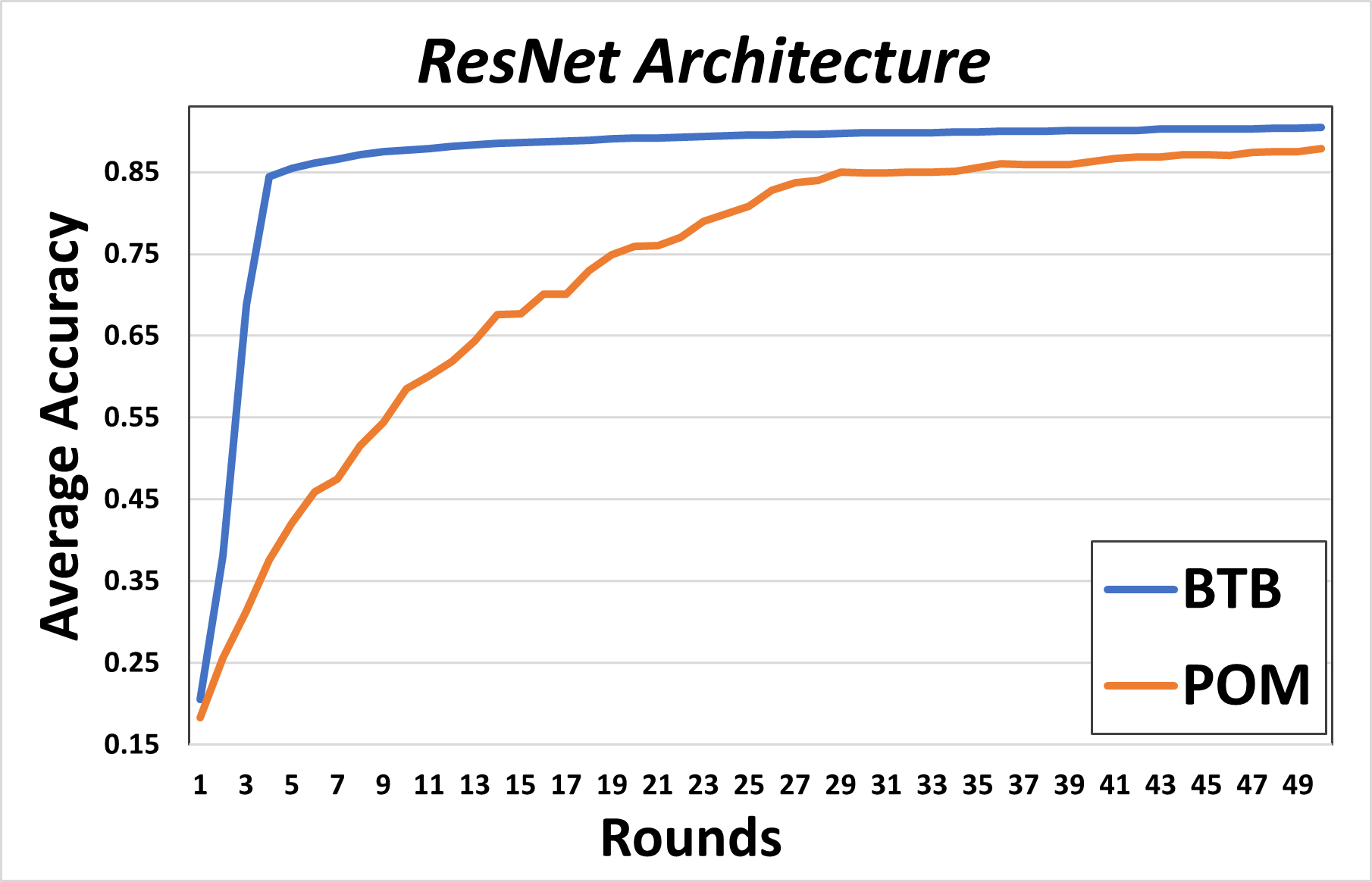}
     \end{subfigure}
        \caption{Comparing {\em POM} with {\em BTB} on ensemble accuracy (upper part), and average learner accuracy (lower part).}
        \label{fig:pom}
\end{figure}

In Figure~\ref{fig:pom} we compare the \textit{POM} with \textit{BTB} which is the best coordinated policy. It can be seen that {\em POM} has a much slower rate of learning relative to the coordinated policy, even in the case of ResNet, where learning is very fast for {\em BTB}. Nevertheless, {\em POM} eventually converges to much higher performance that nearly approaches that of the coordinated policy. Notably, even the average learner accuracy is high. This implies that knowledge diffuses from the Oracle to all individuals. The fact that true knowledge eventually dominates the false and inconsistent information circulating in the population is of independent interest and possibly worthy of further exploration in the context of machine learning.

\noindent $\bullet$ \textit{Coordinated, evaluation-based policies}. We first note that the choice of policy has a negligible on the highest performance levels reached {\em at convergence}. But it is worth noting that even by a small margin, the highest performance was achieved by the {\em BTB} policy in 4 out of the 6 cases, corroborating the insights in~\cite{dong:dygroup}.

\begin{figure}[h]
     \centering
         \includegraphics[width=40mm]{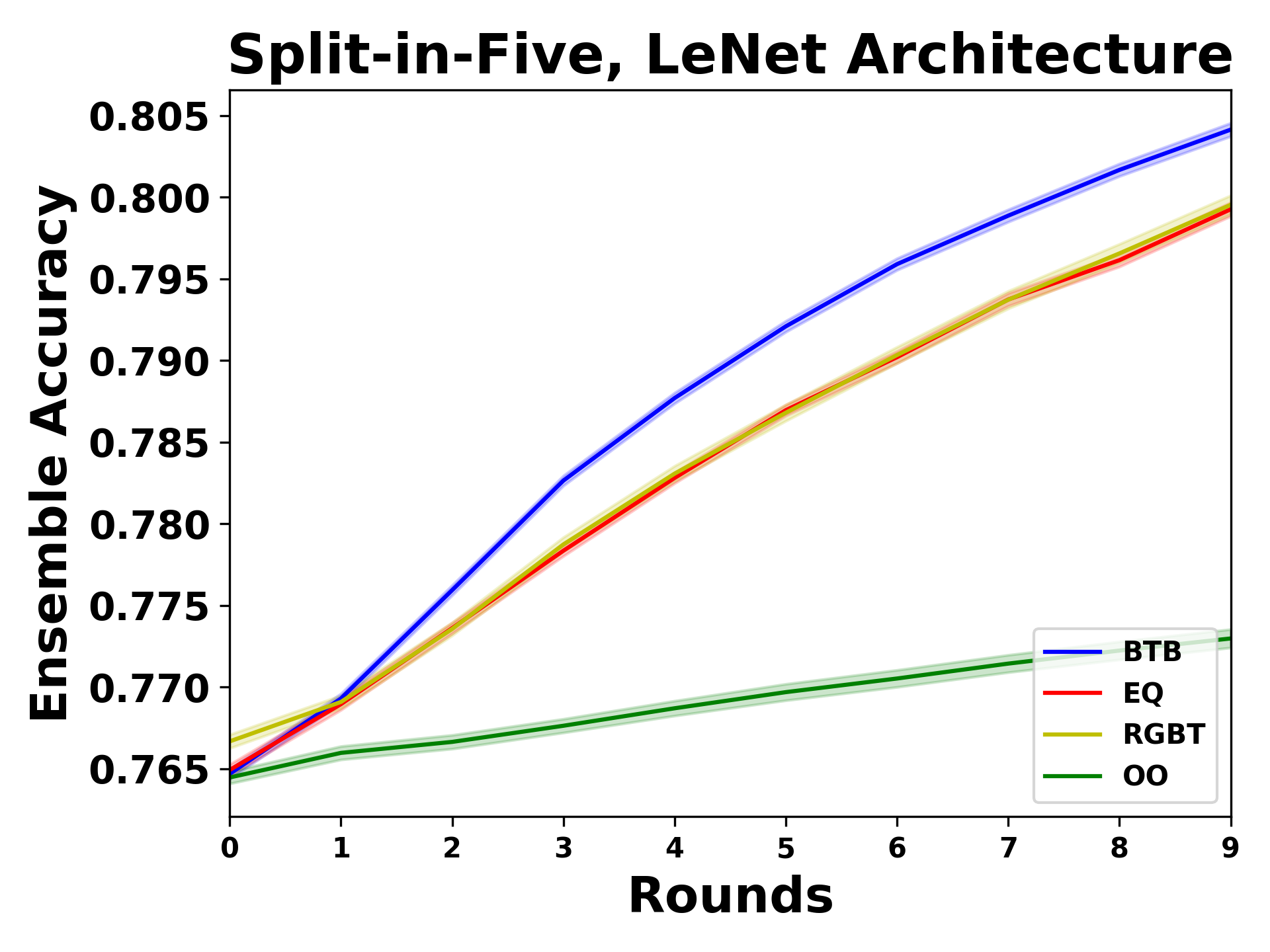}
         \includegraphics[width=40mm]{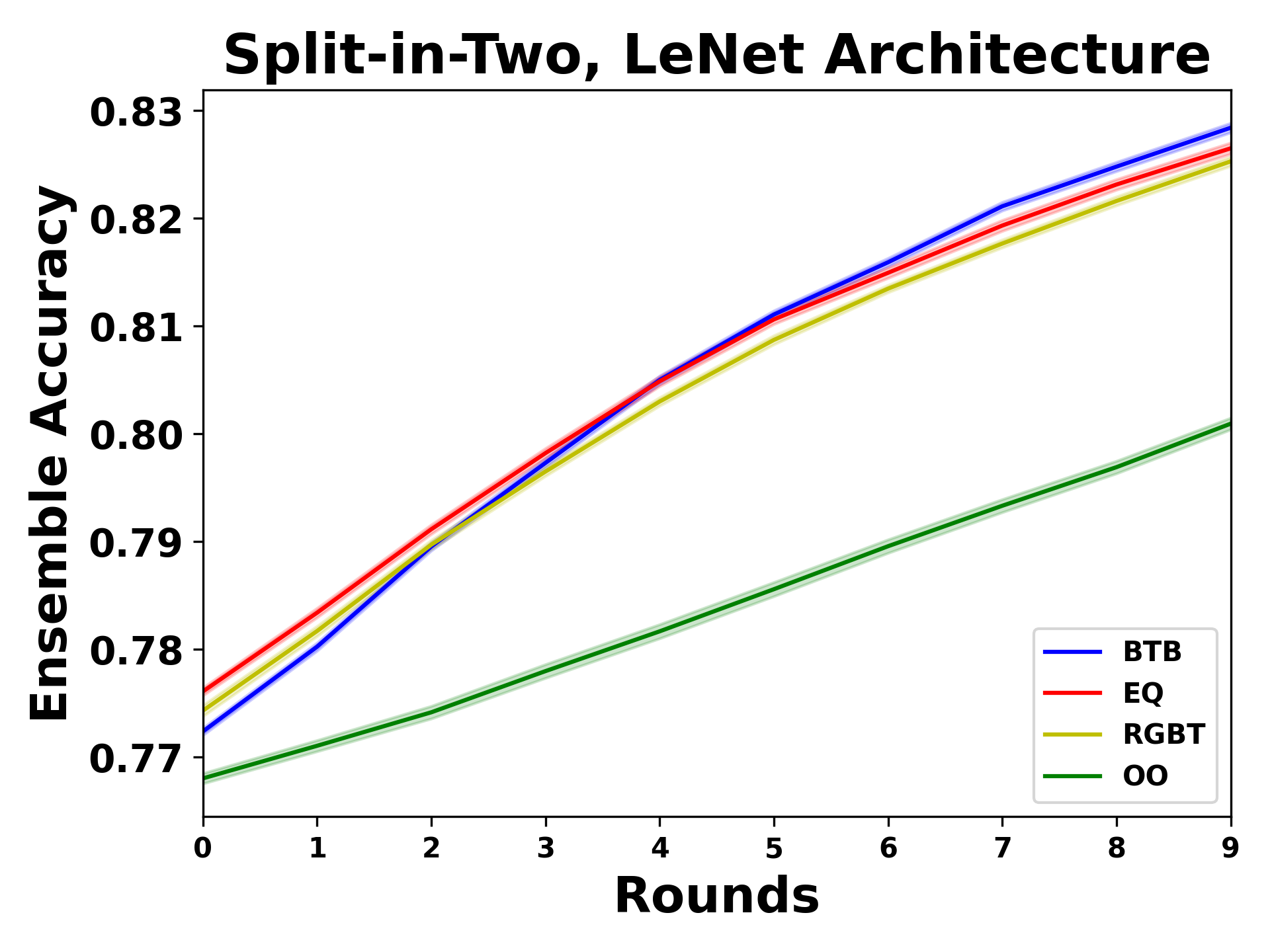}\\
         \includegraphics[width=40mm]{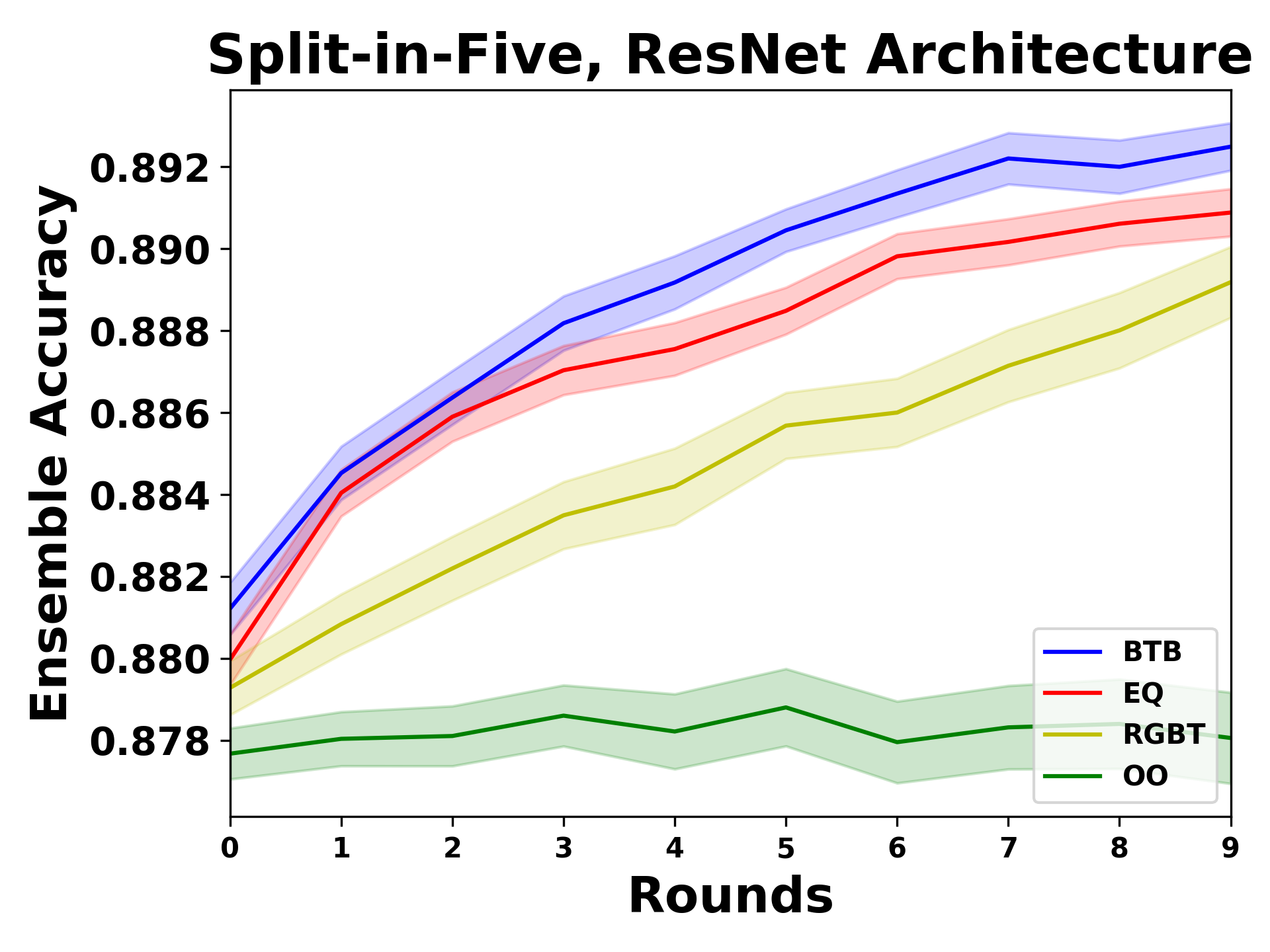}
         \includegraphics[width=40mm]{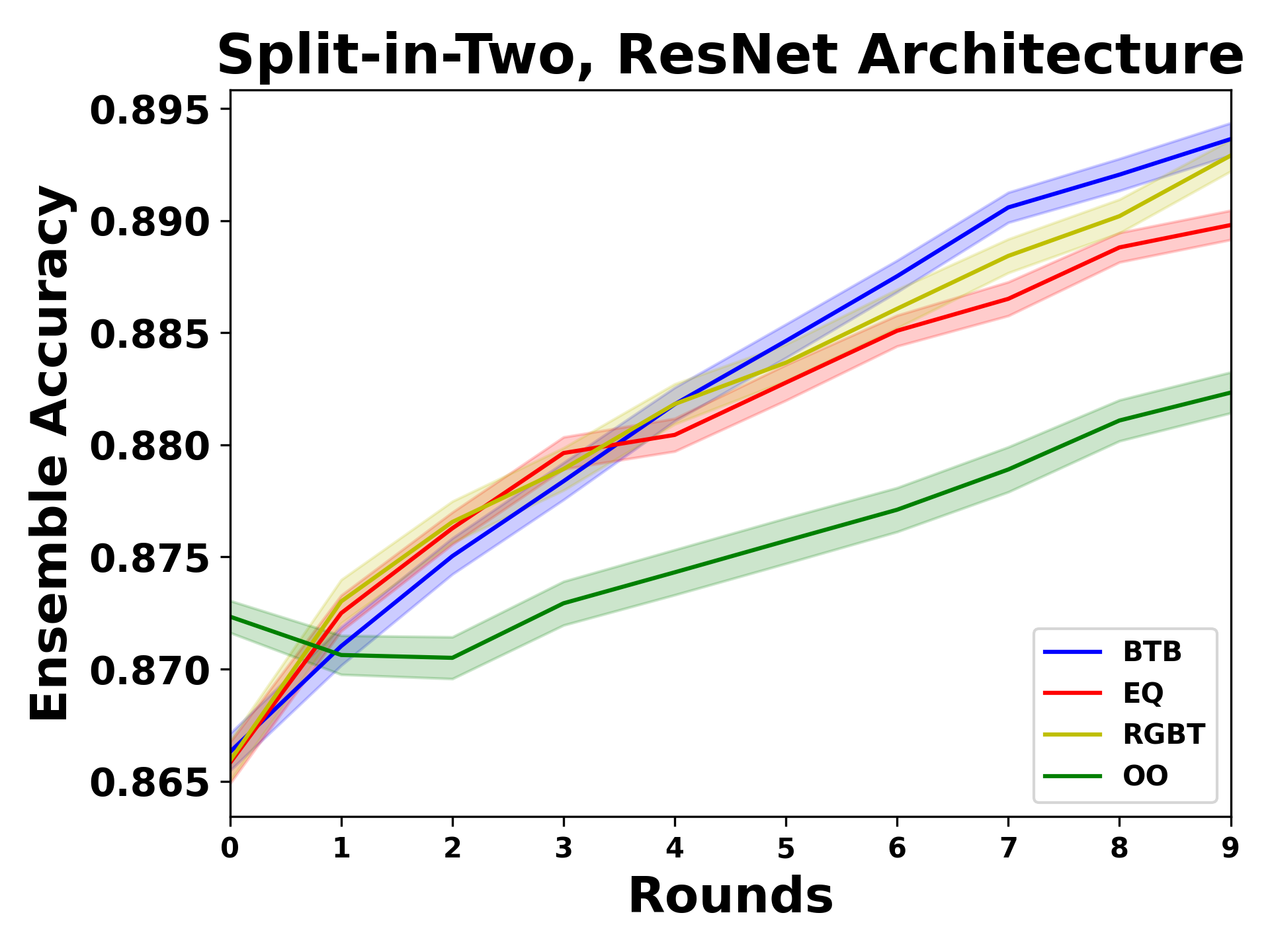}\\
         \includegraphics[width=40mm]{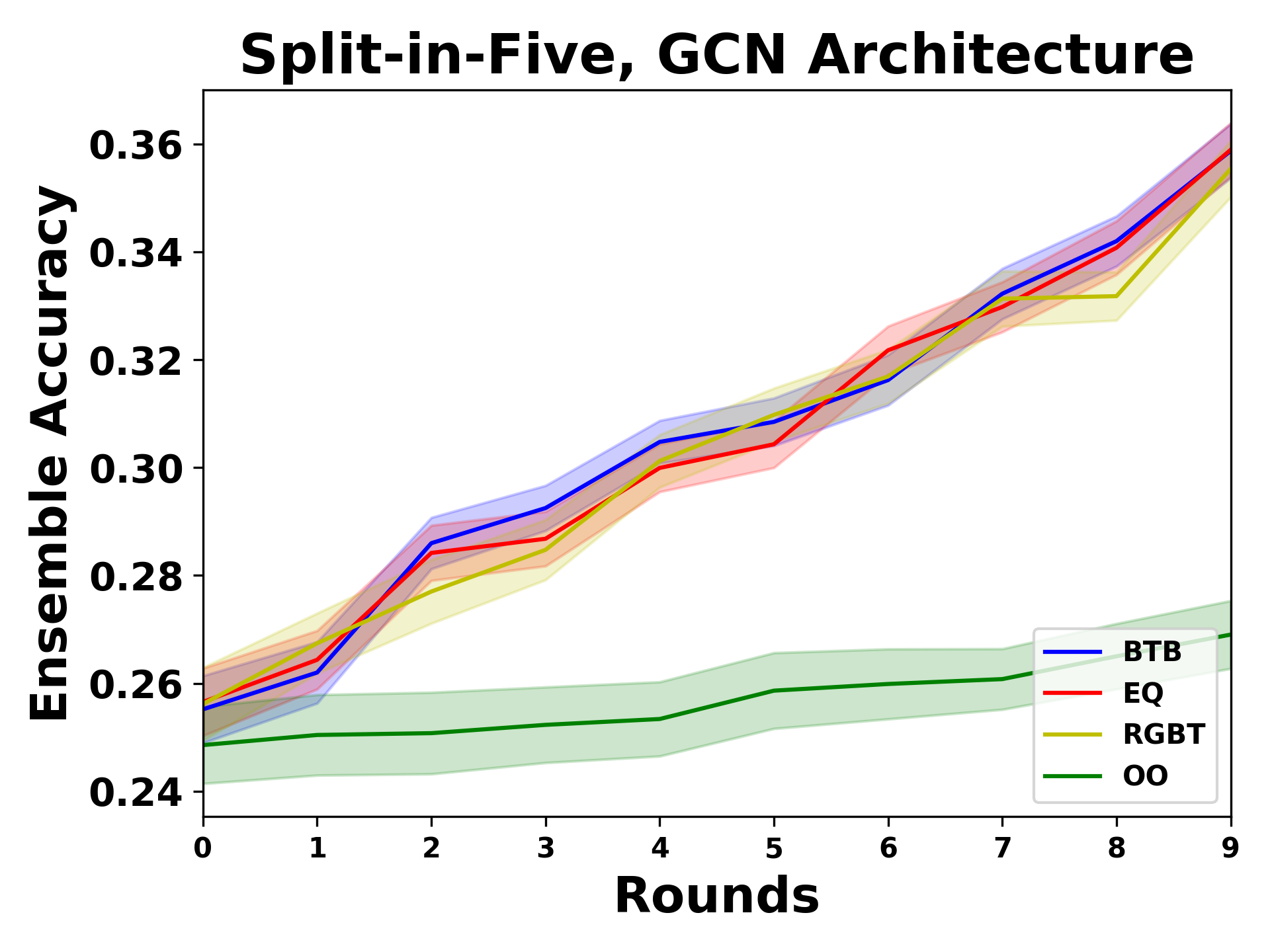}
         \includegraphics[width=40mm]{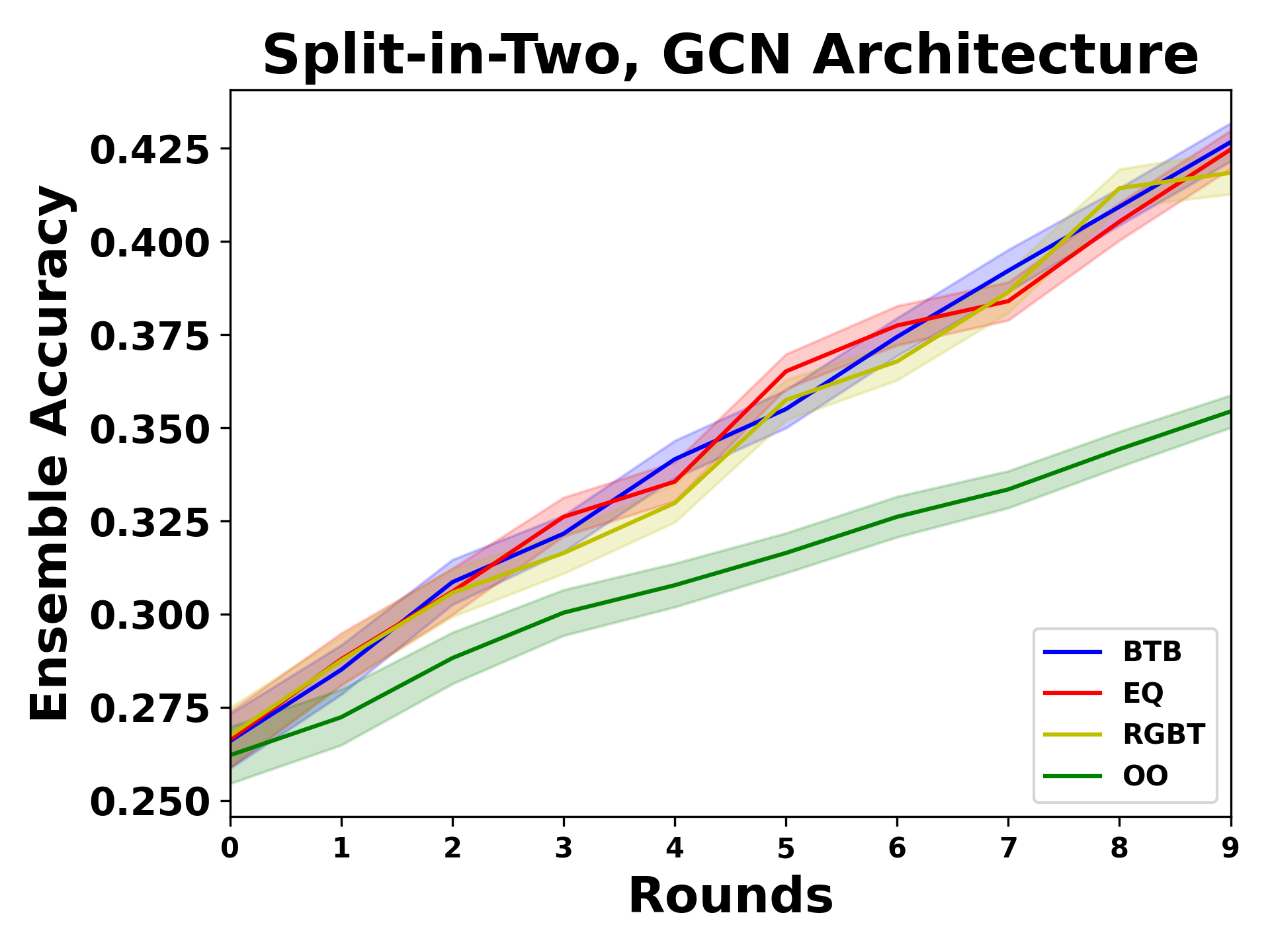} \\
        \caption{Ensemble test accuracy for 10 rounds/epochs with pre-training, for C=2 (left) and C=5 (right). Note that the reported metrics are averages over multiple random experiments, as discussed in Section~\ref{sec:setting}. \textit{BTB} is clearly better on LeNet/ResNet, especially when \textit{C=2}. }
    \label{fig:ensemble_pretrain}
\end{figure}

To our main point in this part, recall that \textsc{nKDiff} is motivated as a framework for training under teaching {\em and} time constraints, favoring faster diffusion, i.e.~better test performance within limited time frames. For that reason we now focus on the first 10 epochs of training and compare the performance of the different policies. Here we use models that have been partially pre-trained at different degrees, as discussed in Section~\ref{sec:formalization}. The results are shown in Figure~\ref{fig:ensemble_pretrain}.

\vspace{-.2cm}

The following points summarize our observations from Figure~\ref{fig:ensemble_pretrain} and other results deferred to the Supplementary Material due to space.\footnote{The Supplementary Material includes a comprehensive set of experiments. In particular, on the topic of policy effect, the differences between policies are less significant without pre-training, although full coordination still appears to be better than moderate coordination.}   

$(a)$ In all cases the average learner accuracy of the {\em Oracle-Only} policy is significantly smaller than all other policies. 
This is also the case for ensemble accuracy, with the exception of LeNet in the no-pretraining case. 
Recall that \textit{OO} makes no use of peer sessions. This indicates that genuine learning takes place in peer sessions.  Thus, if general `educational welfare' metrics are of interest, then peer learning mechanisms appear to be a necessity.

$(b)$ In the case when pre-training is used, there are clear differences between policies. In LeNet and GCN, the fully coordinated policies {\em BTB, EQ} are clearly better than the moderately coordinated {\em RGBT}. The picture is more mixed for ResNet, where in average accuracy, {\em RGBT} does better. This may be due to different properties of the ResNet architecture, or simply to the fact that ResNet learns much more quickly, as noted earlier.

\subsection{Memorization and Robustness to Noisy Labels}
\label{sec:generalization}

\noindent $\bullet$ \textit{Memorization.} The surprising observation that \textsc{nKDiff} works even in the very restricted setting of the Planted Oracle Model leads us to a contrarian question inspired by the work in~\cite{zhang:rethinking}: 
{Can \textsc{nKDiff} train the population to memorize \textit{random} labels?}

\begin{figure}[h]
    \centering
    % \subfigure[]
    \includegraphics[width=0.235\textwidth]{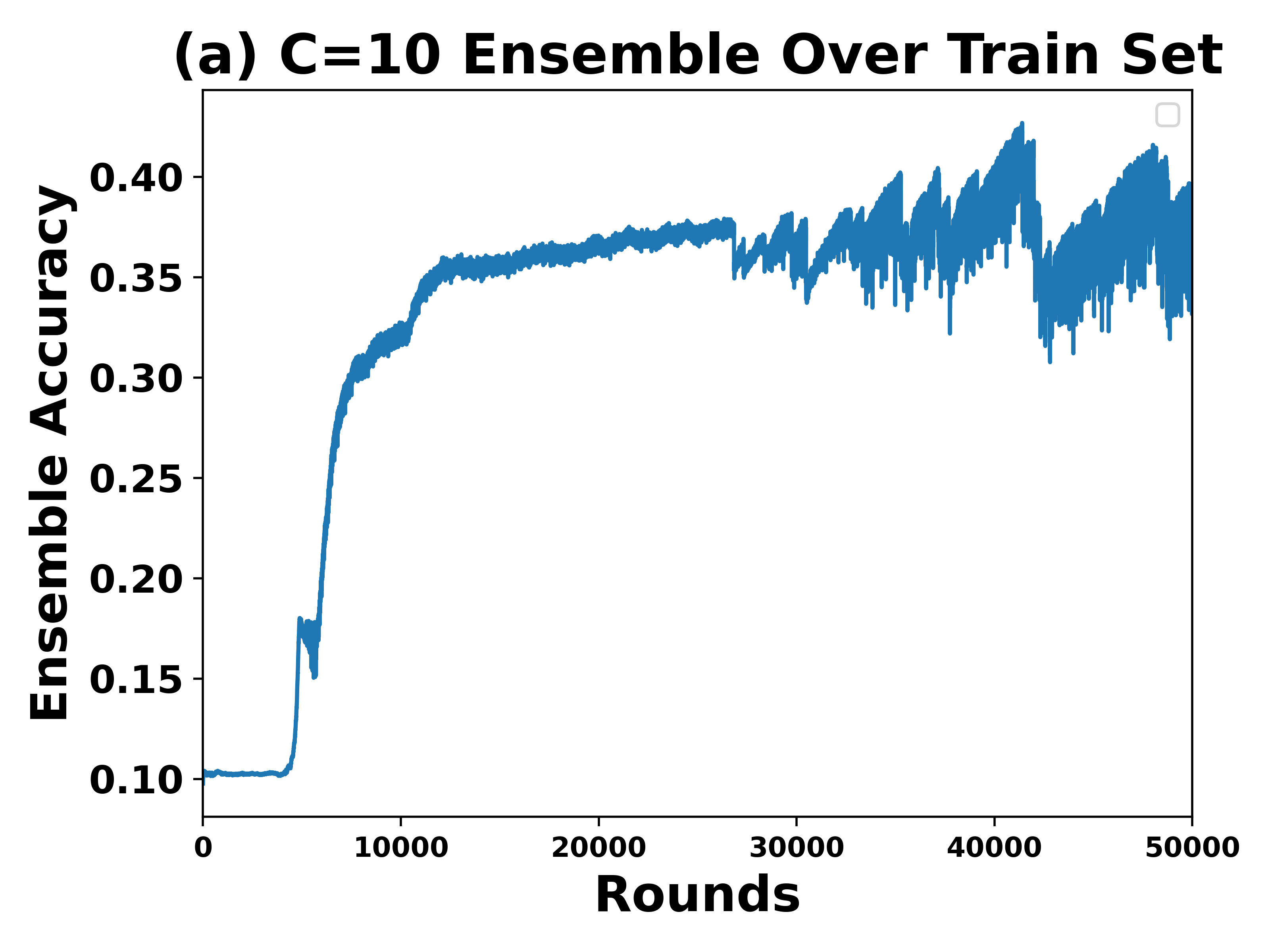}
    \includegraphics[width=0.235\textwidth]{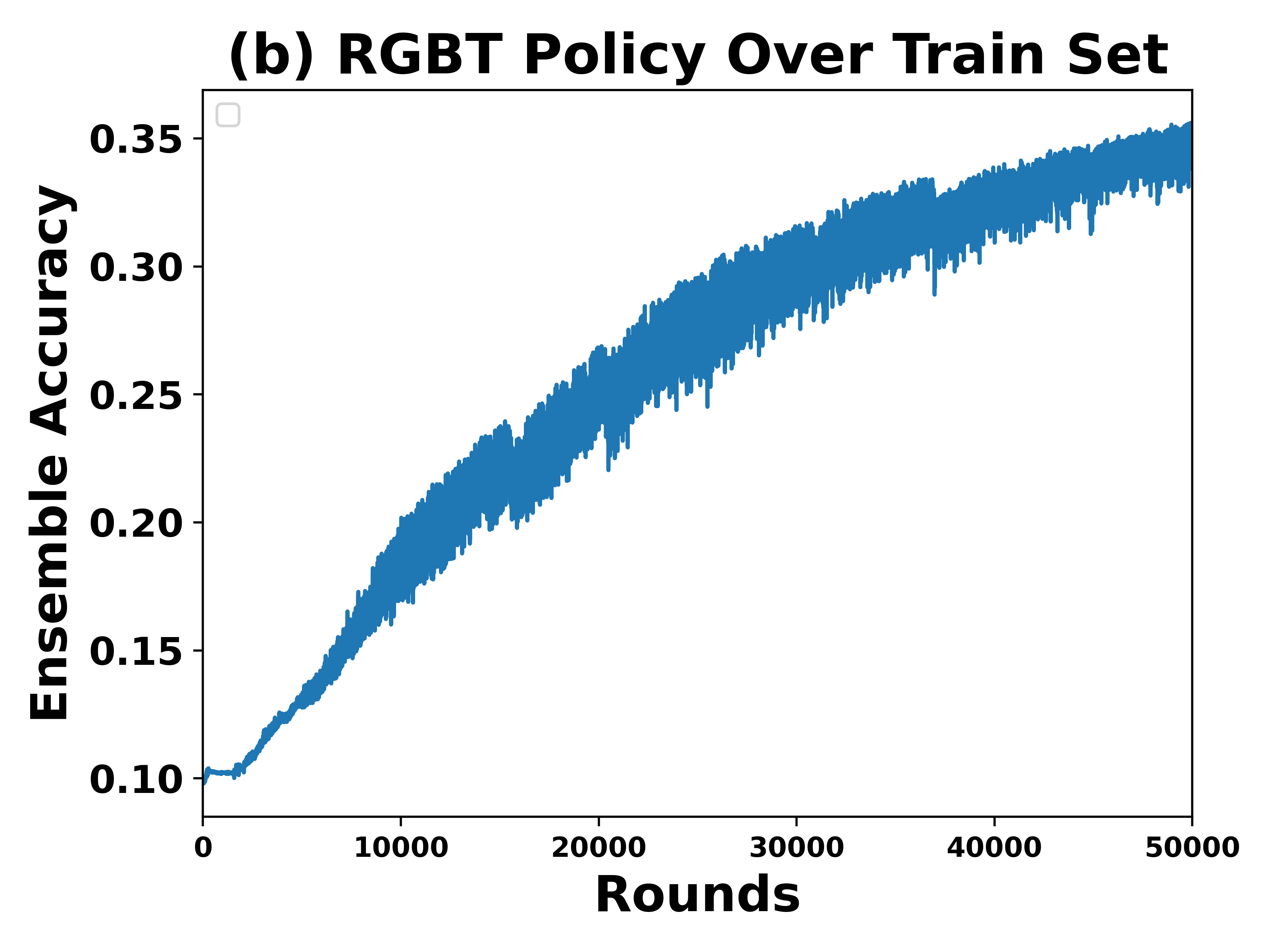} \\
    \includegraphics[width=0.235\textwidth]{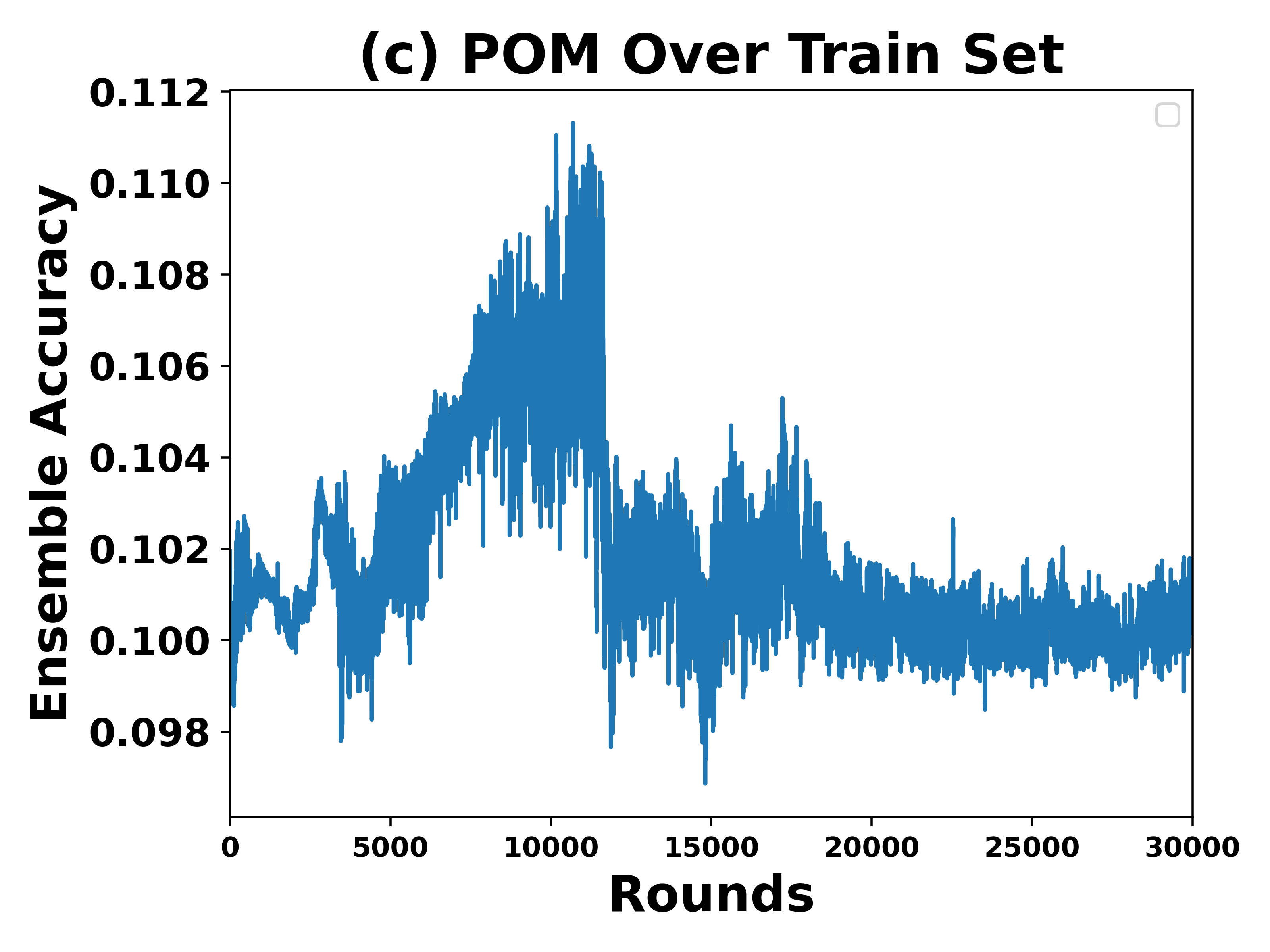}
    \includegraphics[width=0.235\textwidth]{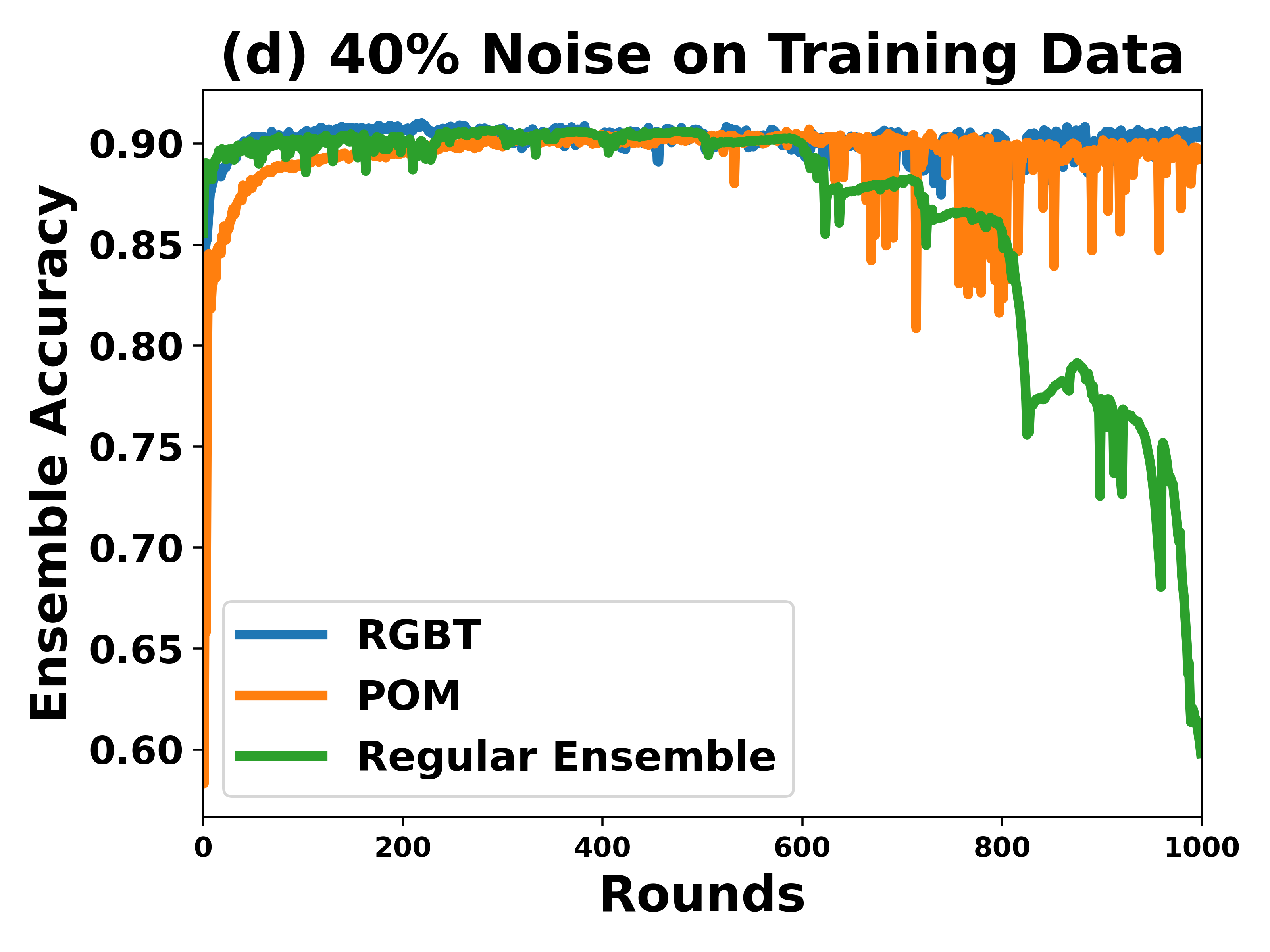}
    \caption{(a,b,c) Training accuracy of LeNet on randomly labeled dataset with 10 classes, for three different \textsc{nKDiff} policies. (d) Test accuracy of ResNet on 40\% noisy training data.}
 \label{fig:memorization}
\end{figure}

\smallskip 
For background, Zhang~et~al.~~\cite{zhang:rethinking} observe that the training accuracy of large deep networks reaches 100\% on training sets with random labels, i.e.~the models have the capacity to fully memorize the data. In our case we use smaller models that still have a significant capacity for memorization after sufficient training. For example, as illustrated in Figure~\ref{fig:memorization}-(a), training the LeNet model on a set $(X,y_r)$ where $y_r$ are random labels of the points in $X$ reaches an accuracy of roughly 40\%, much higher than the random expectation of 10\%. 

% \begin{figure*}[t]
%      \centering
%      \begin{subfigure}[]
%          \centering
%          \includegraphics[width=40mm]{Figures/Memorization/Regular.png}
%          % \caption{\scriptsize \textit{C = 10}}
%      \end{subfigure}
%      \begin{subfigure}[]
%          \centering
%          \includegraphics[width=40mm]{Figures/Memorization/RGBT.png}
%          % \caption{\scriptsize \textit{RGBT}}
%      \end{subfigure}
%      \begin{subfigure}[]
%          \centering
%          \includegraphics[width=40mm]{Figures/Memorization/POM.png}
%          % \caption{\scriptsize \textit{POM}}
%      \end{subfigure}
%      \begin{subfigure}[]
%          \centering
%          \includegraphics[width=40mm]
%          % \includegraphics[width=37mm]
%          {Figures/Soft-Hard_Accuracy/Lenet_soft-hard-1.png}
%          % \caption{\scriptsize \textit{Two LeNets Agreement}}
%      \end{subfigure}
%         \caption{(a,b,c) Training accuracy on randomly labeled dataset with 10 classes, for three different emsemble training policies.\\ (d) Two LeNet models: the size of intersection and union of their correctly classified test points. }
%         \label{fig:memorization}
%         \vspace{-0.3cm}
% \end{figure*}

\vspace{-.1cm}

We next try a moderately coordinated, evaluation-based policy, \textit{RGBT}, for \textit{C=2}. In Figure~\ref{fig:memorization}-(b) we observe that memorization does take place, albeit at a much slower rate, reaching a lower value of 35\%.  
However, when we use the {\em uncoordinated} \textit{POM}, we see in Figure~\ref{fig:memorization}-(c) that training accuracy stays between 10\% and 11\% throughout training. 
In combination with the results in Section~\ref{sec:policyEffect}, this suggests that {\em POM} is able to learn a very effective classifier, which however has no capacity of memorizing training points. In particular, it appears that the somewhat reduced convergence rate and test performance of \textit{POM} comes with the benefit of memorization resistance.\footnote{Up to our knowledge, none of the existing methods for learning from noisy labels~\cite{noisySurvey} have been tested at the extreme of fully random labels. Thus, \textsc{nKDiff} with \textit{POM} may be the only known training algorithm that is resistant to memorization. However, this requires further investigation. }

\smallskip

\noindent $\bullet$ \textit{Robustness to noisy labels.} The next logical step is to consider the more general case when {\em a fraction} of the labels are corrupted randomly, where overfitting can  become a serious issue. Figure~\ref{fig:memorization}-(d) depicts a single experiment with ResNet, over 1000 epochs. In this case, 40\% noise was added to training data.  In our \textsc{nKDiff} framework, \textit{POM} and \textit{RGBT} are {\em robust} against the noisy training data and very little  overfitting happens. In contrast, when all models learn from the (noisy) Oracle, test accuracy dropped almost 30\% due to overfitting which started after 600 epochs. It is also interesting while overfitting episodes happen after 600 epoch for \textit{POM} and \textit{RGBT}, \textsc{nKDiff} exhibits self-correcting behavior, presumably due to the `healty' learners that overpower the noisy Oracle.

\section{Conclusion}
\label{sec:open}

We have initiated a study of \textit{natural} Knowledge Diffusion processes for training populations of artificial learners. The work is  motivated by the problem of peer group formation in human educational systems, a sensitive and widely debated issue in social psychology and policy making. While the analytical modeling or `artificialization' of similar questions may be an interesting line of work that can help researchers develop real-world insights~\cite{dong:dygroup}, we wish to emphasize that it is not our intention to take a stance, make policy recommendations, or claim any impact outside the field of artificial intelligence.

In the context of machine learning, we did not aim to solve an existing problem, but to explore generalization phenomena under natural resource constraints. As discussed in section~\ref{sec:related}, many recent works aim to design sophisticated algorithms for addressing pitfalls of overparameterization, such as model complexity and overfitting to noisy data. In our work, dealing with natural resource constraints led us to explore the possibility that overparameterization, in synergy with stochasticity, is a \textit{natural resource} that enables populations of learners to mine faster, more efficiently, and without overfitting, the ground truth knowledge. Interestingly, that emerges with a higher-order overparameterization at the population level; {\em more} trainable parameters are now distributed to multiple individuals who communicate via simple diffusion processes. 

While we envisioned \textsc{nKDiff} as a population-level process, our individuals are simple identical-architecture neural models that can be alternatively conceived as a single ensemble model with modular characteristics~\cite{Robbins:modularity}. The notion and role of modularity have been extensively investigated in biology, in particular in connection with the architecture of the mind~\cite{Fodor:modularity,Buller:GetOver,clune:evolutionary}. We find it interesting that the study of generalization under {\em natural} constraints led us to the design of a modular system that as a whole does not have the capacity to memorize despite the ability of its constituent parts to do so, possibly suggesting a mechanism that Natural Intelligence uses to generalize without memorizing.

\medskip
Multiple questions arise, including the following: 

$(a).$ We used training rounds that coincide with standard epochs over existing training benchmark datasets. A more detailed picture may emerge for rounds of finer granularity.  Related to that is the effect of population size. In our experiments, we have found that even when learners interact with the true labels only 11\% of their learning time, the population (and -on average- the individual learners) reach high test accuracy. Increasing the size $N$ of the population would enable situations with a lower rate of access to the Oracle. It is then interesting to study whether and under what conditions, interaction with inconsistent labels becomes potentially catastrophic for the performance of peer-trained populations.

$(b).$ Robustness to noisy labels emerged as a byproduct of our study. This phenomenon is worth of further study. An open question is whether knowledge diffusion mechanisms can yield any advantages over existing algorithms for dealing with noisy labels. It would also be interesting to explore whether populations trained with \textsc{nKDiff} are robust to other types of noise and their adaptability to distribution shifts.

More generally, our study revealed various intriguing phenomena. We believe that this framework where {\em `Machine Learning meets Knowledge Diffusion'} opens up multiple directions for future research.

\smallskip
\textbf{Acknowledgements.} Preliminary work was presented at the ICLR 2022 Workshop on Agent Learning in Open-Endedness \cite{beikihassan:aloe22}. This work was supported by and a Faculty Seed Grant from NJIT and NSF Grant \#997421.

\clearpage

\bibliography{references}

\end{document}